\documentclass{article}

\usepackage[utf8]{inputenc} 
\usepackage[T1]{fontenc}    
\usepackage{hyperref}       
\usepackage{url}            
\usepackage{booktabs}       
\usepackage{amsfonts}       
\usepackage{nicefrac}       
\usepackage{color}
\usepackage{xcolor}
\usepackage{graphicx}
\usepackage{floatrow}
\usepackage[percent]{overpic}
\usepackage{amsmath}

\usepackage{algorithm}
\usepackage{algpseudocode}

\usepackage{microtype}
\usepackage{amsmath}
\usepackage{mathtools}
\usepackage{amsfonts}
\usepackage{graphicx}
\usepackage{color}
\usepackage{booktabs} 
\usepackage{wrapfig}
\usepackage{fullpage}

\usepackage{hyperref}

\newcommand{\omt}[1]{}

\def\Prf{{\rm Pr}}

\newcommand{\Prb}[1]{
\Prf\left[{#1}\right]
}

\title{
The Algorithmic Automation Problem: Prediction, Triage, and Human Effort
}

\date{}

\author{
  Maithra Raghu$^{1,2}$\thanks{Correspondence to  \texttt{maithrar@gmail.com}}
  \qquad
  Katy Blumer$^{2}$
  \qquad
  Greg Corrado$^{2}$ \\
  \qquad
  Jon Kleinberg$^{1}$
  \qquad
  Ziad Obermeyer$^{3, 4}$ \qquad Sendhil Mullainathan$^{5}$
  \\
  \\
  $^1$ Department of Computer Science, Cornell University \\
  $^2$ Google Brain \\ 
  $^3$ UC Berkeley School of Public Health\\
  $^4$ Center for Clinical Data Science, MGH \\
  $^5$ Chicago Booth School of Business
}

\begin{document}

\maketitle

\begin{abstract}
In a wide array of areas, algorithms are matching and surpassing
the performance of human experts, leading to 
consideration of the roles of human judgment and algorithmic prediction in these domains.
The discussion around these developments, however, has implicitly
equated the specific task of prediction with the general task of automation.
We argue here that automation is broader than just a comparison of human versus algorithmic performance on a task;
it also involves the decision of which instances of the task to give to the algorithm in the first place.
We develop a general framework that poses this latter
decision as an optimization problem, and we show how 
basic heuristics for this optimization problem can lead
to performance gains even on heavily-studied applications of AI in medicine.
Our framework also serves to highlight how effective automation depends
crucially on estimating
both algorithmic and human error on an instance-by-instance basis, and
our results show how improvements in these error estimation problems
can yield significant gains for automation as well.
\end{abstract}

\section{Introduction}

On a variety of high-stakes tasks, machine learning algorithms are on the threshold of doing what human experts do with such high fidelity that we are contemplating using their predictions as a substitute for human output. For example, convolutional neural networks are close to diagnosing pneumonia from chest X-rays better than radiologists can \cite{rajpurkar2017chexnet, topol2019medicine}; examples like these underpin much of the widespread discussion of algorithmic automation in these tasks. 

In assessing the potential for algorithms, however, the community has implicitly equated the specific task of prediction with the general task of automation. We argue here that this implicit correspondence misses key aspects of the automation problem; a broader conceptualization of automation can lead directly to concrete benefits in some of the key application areas where this process is unfolding.

We start from the premise that automation is more than just the replacement of human effort on a task; it is also the meta-decision of which instances of the task to automate.  And it is here that algorithms distinguish themselves from earlier technology used for automation, because they can actively take part in this decision of what to automate.  But as currently constructed, they are not set up to help with this second part of the problem.  The automation problem, then, should involve an algorithm that on any given instance both (i) produces a prediction output; and (ii) additionally also produces a triage judgment of its effectiveness relative to the human effort it would replace on that instance. 

Viewed in this light, machine learning algorithms as currently constructed only solve the first problem; they do not pose or solve the second problem.  In effect, currently when we contemplate automation using these algorithms, we are implicitly assuming that we will automate all instances or none. In this paper, we argue that when algorithms are built to solve both problems -- prediction and triage -- overall performance is significantly higher. In fact, even on tasks where the algorithm significantly outperforms humans on average per instance, the optimal solution is to automate only a fraction of the instances and to optimally divide up the available human effort on the remaining ones. And correspondingly, even on tasks where an algorithm does not beat human experts, the optimal solution may still be to automate a subset of instances. 

Now is the right time to ask these questions because the AI
community is on the verge of translating some of its most successful
algorithms into clinical practice. Notably, an influential line of
work showed how a well-constructed convolutional net trained
on gold-standard consensus labels for diagnosing diabetic retinopathy
(DR) outperforms ophthalmologists in aggregate, and 
these results have led to considerable optimism about the 
role of algorithms in this setting \cite{topol2019medicine}.
But the community's discussion around these prospects has focused
on the algorithms' per-instance prediction performance without 
considering the problem of recognizing which instances to automate.



Using largely the same data, we build this additional, crucial component and find that, even in this context where an algorithm outperforms human experts in the aggregate, the optimal level of triage is not full automation. Instead significantly more accuracy can be had by triaging a fraction of the instances to the algorithm and leaving the remaining fraction to the human experts.  Specifically, full automation reduces the error rate from roughly 5.5\% by human doctors to 4\% with an algorithm solving every instance; automation with optimal triage, though, reduces the error further to roughly 3.5\% -- effectively adding a significant further fraction to the gains that were realized by algorithmically automating the task in the first place.  

This gain occurs for two reasons: first the algorithm’s high average performance hides significant heterogeneity in performance. For example, on roughly 40\% of the instances the algorithm has {\em zero} errors. By the same token, on a small set of instances, the algorithm makes far more errors than average and these instances can be assigned to humans.  Second, when the algorithm automates a fraction of the cases, that frees up human effort; reallocating that effort to the remaining cases can achieve further gains.  In principle these gains could come from a single doctor spending additional time on the instance, or from multiple doctors looking at it; in our case, the available data allows us to explicitly quantify the gains arising from the second of these effects, due to the fact that we have multiple doctor judgments on each instance. 

These results empirically demonstrate the importance of the triage component for the automation problem. We show that the gains we demonstrate are unlikely to have fully tapped the potential gains to be had through algorithmic triage: this neglected component deserves the kind of sustained effort from the machine learning community that the prediction component has received to date. In fact, given the disparity in efforts on these two problems, it is possible that the highest return to improving automation performance is through solving triage rather than further improving prediction. 

\section{General Framework}
\label{sec:framework}

In a typical application where we consider using algorithmic effort in place of human effort, the goal is to {\em label} a set of instances of
the problem with a positive or a negative label.
For example, in a medical diagnosis setting, we may have a set of medical
images, and the goal is to label each with a binary yes/no diagnosis.
Let $x$ be an instance of the labeling problem, and let 
$a(x)$ be its {\em ground truth} value.
For our purposes (as in the example of a binary diagnosis), 
we will think of this ground truth value as taking a value of 
either $0$ or $1$, 
with $a(x) = 0$ corresponding to a negative label and $a(x) = 1$
corresponding to a positive label.

How do we approach this problem algorithmically?
Given a set $U$ of instances, 
we could train an algorithm to produce a numerical estimate
$m(x) \in [0,1]$ with the goal of minimizing a
loss function $\sum_{x \in U} L(a(x),m(x))$, where $L(\cdot,\cdot)$
increases with the distance between its two inputs.
For notational convenience, we will write $g(x)$ for $L(a(x),m(x))$,
the algorithmic error on instance $x$.
The $m(x)$ values are then converted into (binary) predictions, and we can
evaluate the resulting error relative to ground truth. As a concrete example, one option is to threshold the $m(x)$ to produce a $0$ or $1$ value, and evaluate agreement with $a(x)$.

When a social planner considers the prospect of introducing
algorithms into an existing task, we often imagine the question
to be the following.
The planner currently has human effort being devoted to instances of the task;
for an instance $x \in U$, we can imagine that there is a human output
$h(x)$, resulting in a loss $f(x) = L(a(x),h(x))$.
The question of whether to automate the task could then be viewed
as a comparison between $\sum_{x \in U} g(x)$ and $\sum_{x \in U} f(x)$ ---
the loss from algorithmic effort relative to the loss from human effort.

\paragraph{Allocating Human Effort}
In order to think about the activity of automation in a richer sense,
it is useful to start from the realization that even in the absence
of algorithms, the social planner is implicitly working with a
larger space of choices than the simple picture above suggests.
In particular, they have some available budget of total human effort,
and they do not need to allocate it uniformly across instances:
for an instance $x$, the planner can allocate $k$ units of
human effort for different possible values of $k$. 
There are multiple possible interpretations for the meaning of $k$;
for example, in the case of diagnosis we could think of $k$
as corresponding to the number of distinct doctors who look
at the instance, or alternately to the total amount of time spent collectively
by doctors on the instance.
Thus, our functions $h$ and $f$ should more properly be written
as two-variable functions that take an instance $x$ and a level of effort $k$:
we say that $h(x,k)$ is the 
label provided as a result of $k$ units of human effort on instance $x$,
and $f(x,k) = L(a(x), h(x,k))$ is
the resulting loss that we would like to minimize.

Note that as functions of effort $k$, 
it may be that $f(x,k)$ and $f(x',k)$ are quite different
for different instances $x$ and $x'$.
For example, instance $x$ may be much harder than instance $x'$,
and hence $f(x,k)$ will be much larger than $f(x',k)$;
similarly, instance $x$ might not exhibit as much marginal benefit from
additional effort as instance $x'$, and hence the growth of
$f(x,k)$ over increasing values of $k$ might be much flatter than 
the growth of $f(x',k)$.
The social planner might well not have precise quantitative
estimates for values like $f(x,k)$, but implicitly they are
seeking to allocate human effort across the set of instances $U$
so as to minimize the total loss incurred.
And indeed, a number of basic protocols --- such as asking for
second opinions --- can be viewed as increasing the amount of effort
spent on instances where there might be benefits for error reduction.

\subsection{Automation involving Algorithms and Humans}

When algorithms are introduced, the social planner has several new
considerations to take into account.
First, the full automation
problem should be viewed more broadly than just a binary comparison 
of human and algorithmic performance;
it can involve decisions about the allocation of both
human and algorithmic effort.
The introduction of the algorithm need not be all-or-nothing: 
we can choose to apply it to some instances in a way that replaces 
human effort, thereby potentially freeing up this effort to be used 
on other instances. 
The average overall comparison might even hide instances
where the algorithm much more significantly under- (or out-) performs
the human.
Second, decisions about the allocation of human effort depend
on the function $f(x,k)$, which can be challenging to reason about.
Algorithms can potentially provide assistance in estimating these
quantities $f(x,k)$ to help even in the allocation of human effort.

The general problem can therefore be viewed as follows.
We would like to select a subset $S$ of instances on which 
no human effort will be used (only the algorithm), and we
will then optimally allocate human effort on the remaining
instances $T = U - S$.
Suppose that we have a budget $B$ on the total number of units of human
effort that we can allocate, and we decide to allocate 
$k_x$ units of effort to each instance $x \in T$.
On such an instance $x$, we incur a loss of $f(x,k_x)$,
using our notation above; 
and on the instances $x \in S$ we incur a loss of $g(x)$ from
the algorithmic prediction.

We thus have the following optimization problem.
\begin{eqnarray}
& {\rm Min}  & \sum_{x \in S} g(x) + \sum_{x \in T} f(x,k_x) \label {eq:opt} \\
& {\rm subject~to}  & \sum_{x \in T} k_x \leq B \\
& & S \cup T = U; ~~ S \cap T = \phi
\end{eqnarray}

Our earlier discussions about algorithms and humans in
isolation are special cases of this optimization problem:
{\em full automation}, when the algorithm substitutes completely for
human effort, is the case in which $S = U$;
and the social planner's problem in the absence of an algorithm ---
which still involves decisions about the effort variables $k_x$ ---
is the case in which $T = U$.
Intermediate solutions can be viewed as performing a kind of 
{\em triage}, a term we use here in a general sense for a process in which some instances go purely to an algorithm
and others receive human attention.

By deliberately adopting a very general formulation, we can also
get a clearer picture of the kinds of information we would need
in order to perform automation more effectively.
Specifically, 
\begin{itemize}
\item[(i)] In addition to making algorithmic predictions $m(x)$,
the automation problem benefits from more accurate estimates
of the algorithm's instance-specific error rate $g(x)$.
\item[(ii)] The allocation of human effort benefits from better
models of human error rate, including error as a function of
effort spent $f(x,k)$.  As noted above, we can use an algorithm to
help in estimating this human error rate.
\item[(iii)] Given estimates for the functions $f$ and $g$, 
we can obtain further performance improvements purely through
better allocations of human effort in the optimization problem
(\ref{eq:opt}).
\end{itemize}

We note that the notion of human error involves an additional set of
complex design choices,
which is how humans decide to make use of
algorithmic assistance on the instances (in the set $T$) where
they spend effort.  In particular, if we imagine that algorithmic
predictions are available on the instances in $T$, then the humans
involved in the decision on $x \in T$ may have the ability to incorporate
the algorithmic prediction $m(x)$ into 
their overall output $h(x,k_x)$, and this will have
an effect on the error rate $f(x,k_x)$.
In general, of course, it will be difficult to model a priori how
this synthesis will work, although it is a very interesting question; 
we show that our results for the automation problem do not require
assumptions about this aspect of the process, but we explore this
question later in the paper.

\subsection{Heuristics for Automation}

If we think of the social planner as the entity tasked with
solving the automation problem in (\ref{eq:opt}),
they are now faced with a set of considerations:
not simply the binary question of whether to use human or algorithmic 
effort, but instead how to divide the instances between those
(in $S$) that will be fully automated and those (in $T$)
that will involve human effort, and how to estimate the error rates
$g(x)$ and $f(x,k)$ so as to solve the allocation problem effectively.

We will show that significant performance gains can be achieved
over both algorithmic and human effort even if we use only very
simple heuristics for the different components of the allocation problem.
Moreover, through a stronger benchmark based on ground truth,
we will also show that much stronger gains are in principle achievable
with improved approaches to the components.

We can describe the simplest level of heuristics in terms of
sub-problems (i), (ii), and (iii) from earlier in this section. The simplest 
heuristic for (i) is to use the functional form of the variance, $m(x)(1 - m(x))$ as a measure of the algorithm's uncertainty
in its prediction on $x$.
A comparably natural predictor does not exist for (ii). We therefore design new
algorithmic predictors to estimate the values of both (i) and (ii), and use these to guide the allocation of algorithmic and human effort. We show that using separate predictors in this way also strengthens the performance gains relative to the simpler heuristic based on $m(x)(1 - m(x))$, although even this basic heuristic yields improvements over full automation. 

Given these predictors for (i) and (ii), what does this suggest
about simple strategies for approximating
(iii), the allocation of human effort?
First, we could restrict attention to solutions in which 
each instance in $T$ receives the same amount of effort.
Thus, if there are $N$ total instances in $U$, we could
choose a real number $\alpha \in [0,1]$, perform
full automation on a set $S$ of $\alpha N$ instances,
and divide the $B$ units of human effort evenly across the
remaining $\beta = 1 - \alpha$ fraction of the instances.
This means that each instance in the set receiving human effort
gets $B / \beta N$ units of effort.
For simplicity, let us write $B = c N$, so that the human effort
per instance in this remaining set is $c / \beta$.
With this allocation of effort, the resulting loss is 
$\sum_{x \in S} g(x) + \sum_{x \in T} f(x,c/\beta).$

This restriction on the set of possible solutions suggests
the following heuristic.
Consider any partition of the instances into $S$ and $T$,
and suppose we use the algorithm on all the instances.
Then we can write the resulting loss
in the following convenient way:
$\sum_{x \in S} g(x) + \sum_{x \in T} g(x).$
Subtracting from the loss that results when we assign
$c/\beta$ units of human effort to each instance in $T$,
we see that the difference is
$\sum_{x \in T} [f(x,c/\beta) - g(x)]$.

Thus, for a given value of $\alpha$ (specifying the fraction
of instances that we wish to assign to the algorithm),
it is sufficient to rank
all instances $x \in U$ by 
$\tau_\alpha(x) = f(x,c/\beta) - g(x)$,
and then choose the $\alpha N$ instances with 
the largest values of $\tau_\alpha$ to put in the set $S$
that we give to the algorithm.
We can thus think of $\tau_\alpha(x)$ as the 
{\em triage score} of instance $x$,
since it tells us the effect of algorithmic triage relative to human effort
on the expected error.

\subsection{Overview of Results}

We put these ideas together in the context of
a widely studied medical application, concerned with 
the diagnosis of diabetic retinopathy, detailed in the next section.
We rank instances by their triage
score, using simple forms for the algorithmic loss $g(x)$
and human loss $f(x,c/\beta)$, and we then search over possible
values of $\alpha$, evaluating the performance at each.
We find that there is range of values of $\alpha$, and a way of
choosing $\alpha N$ instances to give to the algorithm, so that
the resulting performance exceeds either of the binary choices
of fully assigning the instances to the algorithm or to human effort.

As a scoping exercise, to see how strong the possible gains from
our automation approach might be, we consider what would happen
if we ranked instances by a triage score derived from a
ground-truth estimate of the individual human error on each instance.
Such a benchmark indicates the power of the optimization framework
if we are able to get better approximations to the key
quantities of interest --- the functions $f$ and $g$.
We find large performance gains from this benchmark, and we also
explore some stronger methods to work on closing the gap
between our simple heuristics and this ideal.

\paragraph{Different Costs for Error.}
In many settings, a social planner may associate higher costs
to errors committed by automated methods relative to 
errors committed by humans --- for example, there may be
concern about the difficulty in identifying and correcting
errors through automation, or the end users of the results
may have a preference for human output.
It is natural, therefore, to consider a version of the optimization
problem in which the objective (\ref{eq:opt}) has an additional
parameter $\lambda$ specifying the relative cost of error
between algorithms and humans.
This new objective function is
\begin{equation}
\lambda \sum_{x \in S} g(x) + \sum_{x \in T} f(x,k_x).
\label{eq:opt-lambda}
\end{equation}

One might suppose that as $\lambda$ grows large, the social planner
would tend to favor purely human effort, given the relative cost
of errors from automation.
And indeed, the basic comparison that is typically made between
$\sum_{x \in U} g(x)$ (for full automation) and 
$\sum_{x \in U} f(x,k_x)$ (for purely human effort) would 
suggest that this should be the case, since eventually $\lambda$
will exceed the ratio between these two quantities.
But our more detailed framework makes clear that these
aggregate measures of performance can obscure large levels of
variability in difficulty across instances.
And what we find in our application is that it is possible for
the algorithm to identify 
a large set of instances $S$ on which it makes {\em zero errors}.
Thus, even with strong preferences for human effort over algorithmic
effort, it may still be possible to find sizeable subsets of the
problem that should nevertheless be automated --- a fact that is
hidden by comparisons based purely on aggregate measures of performance.

\section{Medical Preliminaries, Data and Experimental Setup}
We first outline the details of the medical prediction problems, and describe the data and experimental setup used to design the automated decision making algorithm. As our primary goal is to study the interaction of this algorithm with human experts, we treat many of the underlying algorithmic components (e.g. a deep neural network model trained for predictions) as fixed, and focus on the different modes of interactions. 

The main setting for our study is the use of \textit{fundus photographs},
large images of the back of the eye, to automatically detect
\textit{Diabetic Retinopathy} (DR). Diabetic Retinopathy is an eye
disease caused by high blood sugar levels damaging blood vessels in
the retina. It can develop in anyone with diabetes (type 1 or type 2),
and despite being treatable if caught early enough, it remains a
leading cause of blindness \cite{ahsan2015dr}.

\begin{figure}
  \centering
      \begin{tabular}{ccc}
    \hspace*{-10mm} \includegraphics[width=0.2\linewidth]{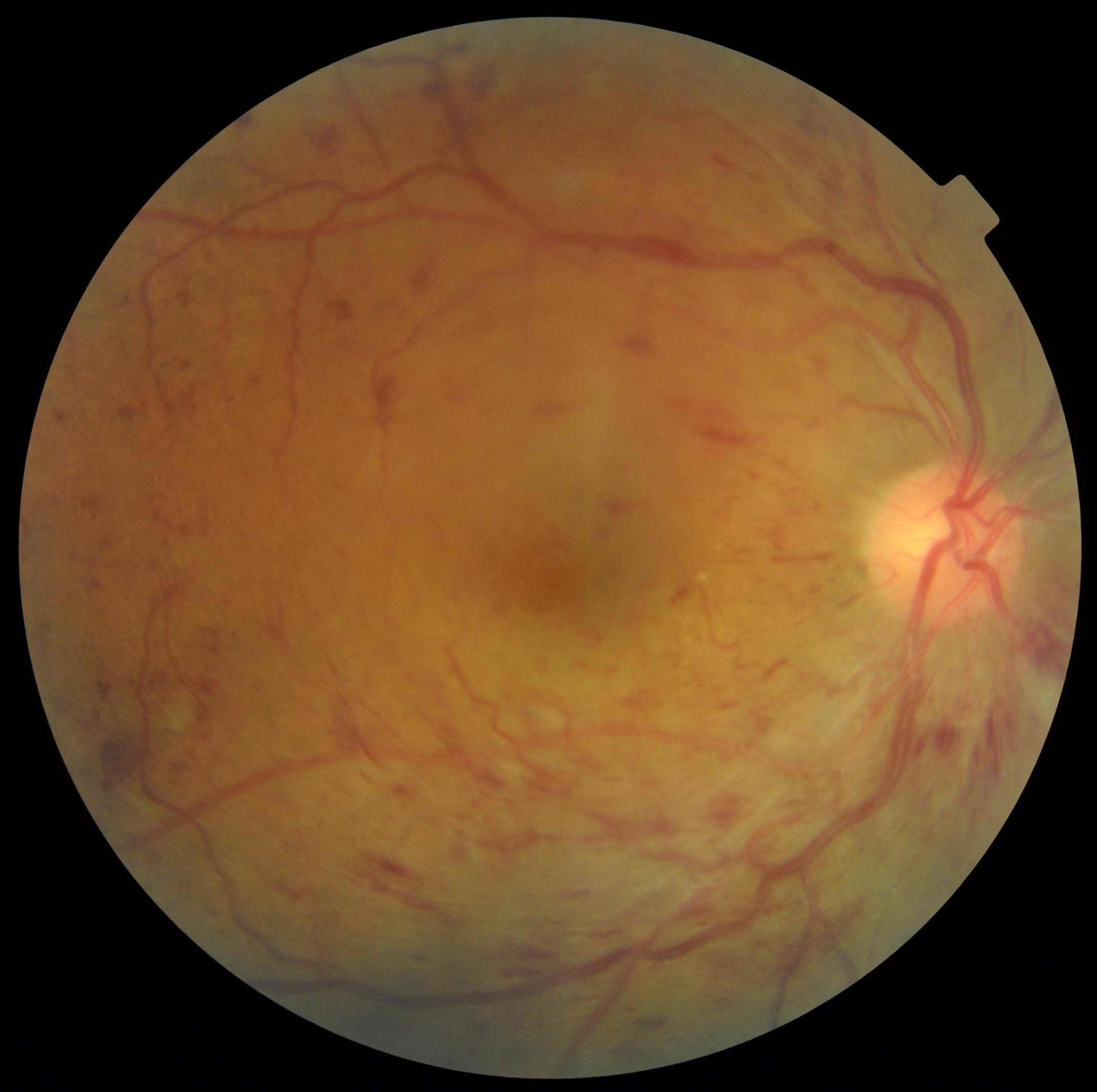} 
    &
    \hspace*{5mm} \includegraphics[width=0.205\linewidth]{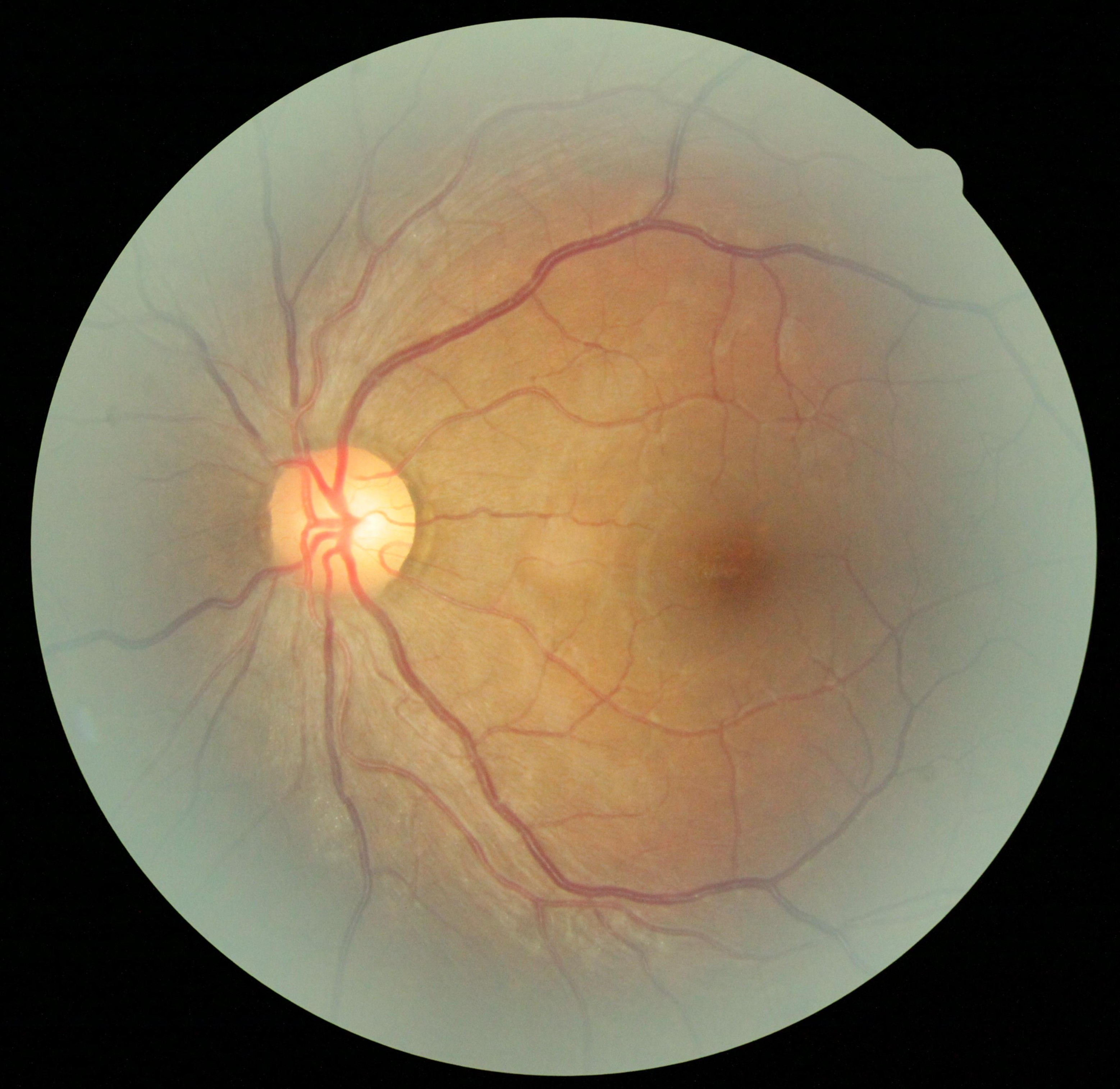}
    &
    \hspace*{5mm} \includegraphics[width=0.25\linewidth]{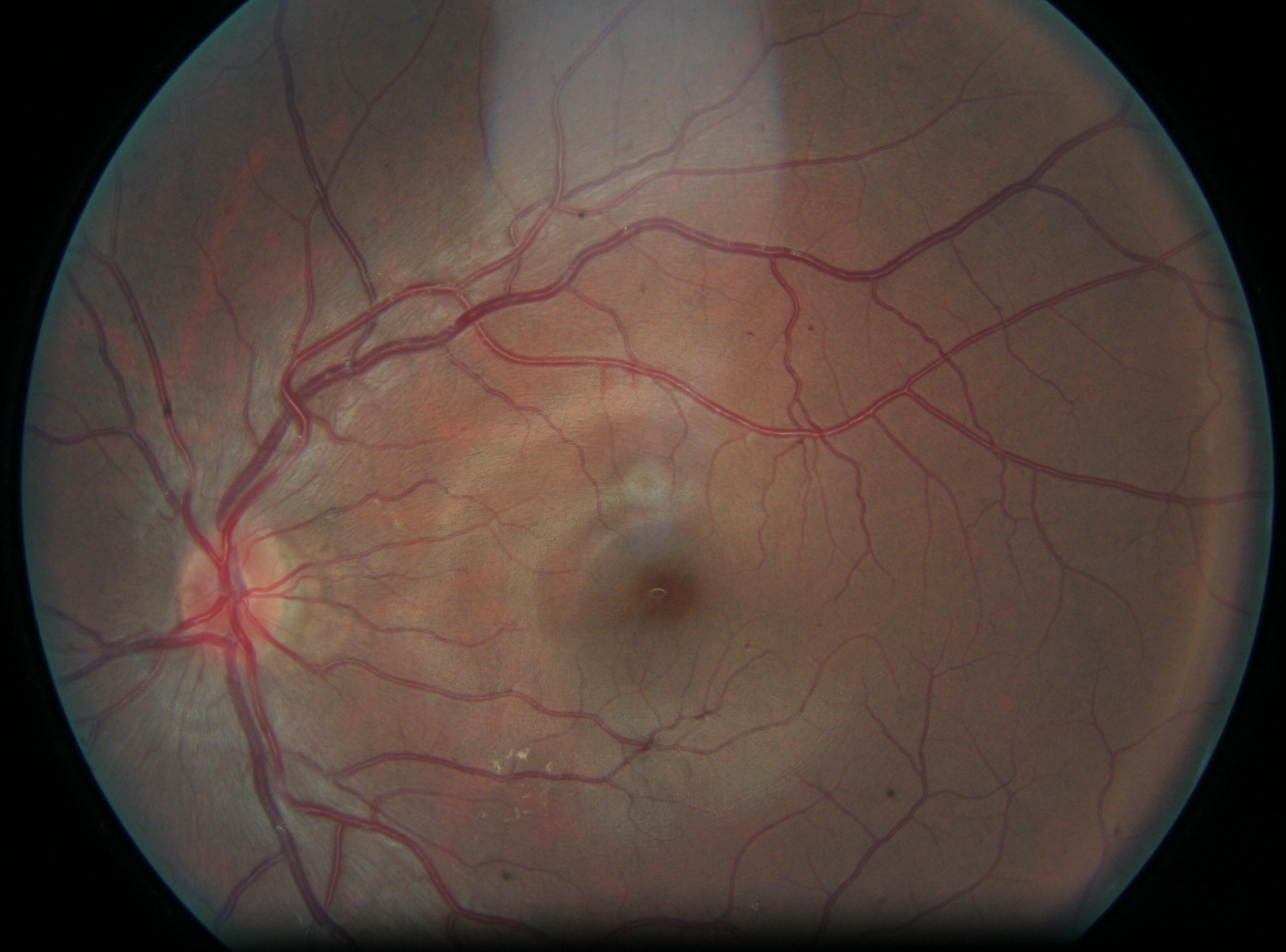}
    \end{tabular}
 \caption{\small \textbf{Example fundus photographs}. Fundus photographs are images of the back of the eye, which can be used by an opthalmologist to diagnose patients with different kinds of eye diseases. One common such eye disease is \textit{Diabetic Retinopathy}, where high blood sugar levels cause damage to blood vessels in the eye.}
 \label{fig-fundus}
\end{figure}

A patient's fundus photograph is graded on a five class scale to
indicate the presence and severity of DR. Grade $1$ corresponds to no
DR, $2$ to mild (nonproliferative) DR, $3$ to moderate
(nonproliferative) DR, $4$ to severe (nonproliferative) DR and $5$ to
proliferative DR. An important clinical threshold is at grade $3$,
with grades $3$ and above called \textit{referable} DR, requiring
immediate specialist attention, \cite{ICDRStandards}. Figure
\ref{fig-fundus} shows some example fundus photos.

\subsection{Data}
The data used for designing the algorithm consists of these fundus photographs, with each photograph having multiple DR grades. These grades are assigned by individual doctors independently looking at the fundus photograph and deciding what DR classification the image should get. There are important distinctions between the data used for training the algorithm, and the data used for evaluation. The training dataset is much larger in size (as a key component is a large deep neural network) and hence each image is more sparsely labelled -- typically with one to three DR grades. The evaluation dataset is much smaller and more extensively annotated. It is described in detail below in Section \ref{sec-evaluation}.

In the mechanics of training our classifier, it will be useful
to view DR diagnosis as a 5-class classification, using the 5-point
grading scheme.
However, when we consider the problem of triage and automation at
a higher level, we will treat the task as a binary classification
problem into images that are referable or non-referable. 



\subsection{A Decision Making Algorithm for Diabetic Retinopathy}
\begin{figure}
\vspace*{-3mm}
  \centering
    \includegraphics[width=0.8\linewidth]{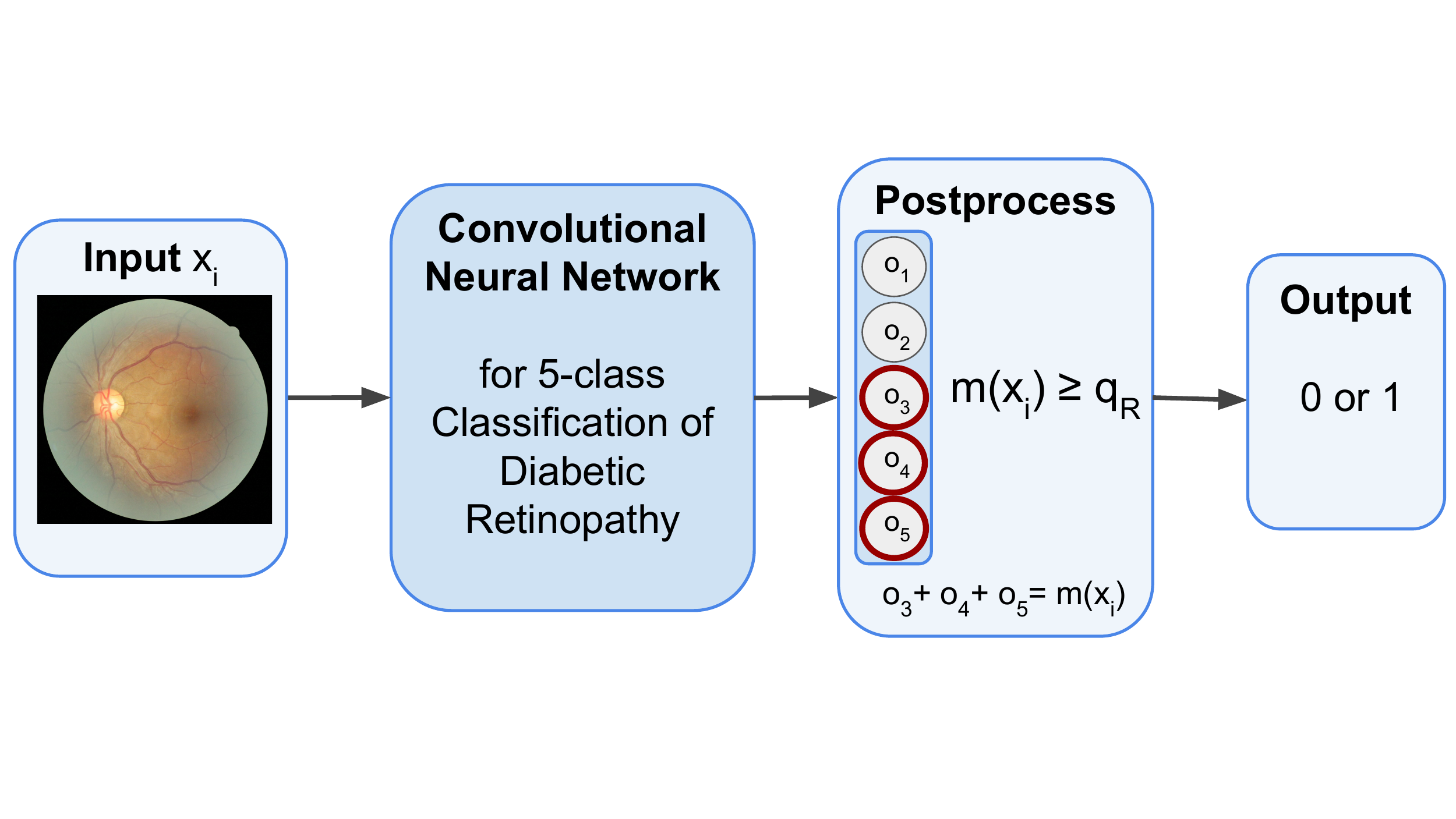}
    \vspace*{-9mm}
 \caption{\small \textbf{Diagram of Algorithm for Diagnosing Diabetic Retinopathy (DR).} The algorithm takes as input a fundus photograph, which, with doctor grades as targets, is used to train a convolutional neural network to perform $5$ class classification of DR. For evaluation on an image $i$ the output values of the convolutional neural network on grades $\geq 3$, $o_3, o_4, o_5$, are summed to give $m(x_i)$, the total output mass on a referable diagnosis. $m(x_i)$ is then thresholded with $q_R$ -- the threshold for a referable diagnosis. This binary decision is output by the algorithm.}
 \label{fig-algorithm}
\end{figure}
Similar to prior work \cite{Gulshan2016Retinal}, we first use the training dataset to train a convolutional neural network to classify each image. Specifically, the CNN outputs a distribution over the $5$ different DR grades for each fundus photograph, with the empirical distribution of the individual doctor grades for that image as the target.  

The question of whether the patient has referable DR (with a grade
of at least 3), and hence needs specialist attention, is one of the
most important clinical decisions.
The outputs of the
trained convolutional neural network form the basis of an algorithm to
make this decision. First, we compute the predicted probability of
referable DR by summing the model's output mass on DR grades $\geq 3$.
For each image $x_i$, this gives a predicted referable DR probability
of $m(x_i)$. Next, we rank the images according to the $m(x_i)$ values, 
and pick a
threshold $q_R$. Images $x_i$ with $m(x_i) \geq q_R$ are labelled as referable
DR by the algorithm, and the others as non-referable.

The choice of the threshold $q_R$ is made so that the total number of
cases marked as referable by the algorithm matches the total number of
cases marked as referable when aggregating the human doctor grades.
This ensures that the effort, resources, and expense needed to act
upon the algorithmic decisions match the current (feasible) effort
resulting from the human decision making process. This is discussed in
further detail in Section \ref{sec-aggregation}

The result of this process is an algorithm taking as input a patient's
fundus photograph, and outputting a binary $0/1$ decision on whether
the patient has non-referable/referable Diabetic Retinopathy. We
illustrate the components of the DR algorithm in Figure
\ref{fig-algorithm}. The full details of our algorithm development setup can be found in Appendix Section \ref{app-sec-train-model}.

\subsection{Evaluation}
\label{sec-evaluation}

We evaluate our decision-making algorithm on a special, gold-standard
\textit{adjudicated} dataset \cite{krause2018adj}. This dataset is
much smaller than our training data, but is meticulously labelled. For
every fundus photograph in the dataset, there are many individual
doctor grades, and also a single \textit{adjudicated} grade, given
after multiple doctors discuss the appropriate diagnosis for the
image. This adjudicated grade acts as a proxy for the ground truth
condition, and we use it to evaluate both the individual human doctors
and the decision making algorithm. In Appendix Section \ref{app-sec-other-dataset} we carry
out an additional evaluation of the methods on a different dataset,
which exhibits the same results.

\subsection{Aggregation and Thresholding}
\label{sec-aggregation}

During evaluation and the triage process, we often have multiple
(binary) grades per image. These grades might correspond to multiple different human doctors individually diagnosing the image, or the algorithm's binary decision along with human doctor grades. In all of these cases, for evaluation, we must typically aggregate these multiple grades into a concrete decision -- a single summary binary grade. To do so, we compute the mean grade and threshold by a value $R$.  If the mean is greater than
$R$, this corresponds to a decision of $1$ (referable); otherwise the
decision is $0$ (non-referable).

The choice of the threshold $R$ also affects the choice of $q_R$ which
is used for the algorithm's decision. To compute $q_R$, we first
aggregate the multiple doctor grades per image into a single grade by
computing their mean and thresholding with $R$. This gives us the
total number of patients marked as referable by the human doctors, and
we pick $q_R$ so that the algorithm matches this number.

In the main text, we give results for $R=0.5$, which corresponds to
the \textit{majority vote} of the multiple grades for an image. In the
Appendix, we include results for $R=0.3, 0.4$, which support the same
conclusions.

\section{The Triage Problem and Human Effort Reallocation}

The performance of human experts and algorithmic decisions are
typically summarized and compared via a single number, such as average
error, F1 score, or AUC. Seeing the algorithm outperform human
experts according to these metrics might suggest the hypothesis that
the algorithm uniformly outperforms human experts on any given
instance.

What we find instead, however, is significant diversity across
instances in the performance of humans and algorithms: 
for natural definitions of human and algorithmic error probability
(formalized below), there are instances in which human effort has lower
error probability than the algorithm, and instances in which 
the algorithm has lower error probability than human effort.
Moreover, this diversity is partially {\em predictable}: we
can identify with non-trivial accuracy those instances on which
one entity or the other will do better.
This diversity and its predictability is an important component of
the automation framework, since it makes it possible to divide
instances between algorithms and humans so that each party is working
on those instances that are best suited to it.

We first study this performance diversity, and then move on to
the problem of allocating effort between humans and algorithms
across instances.

\subsection{Per Instance Error Diversity of Humans and Algorithms}
\begin{figure}
  \centering
      \begin{tabular}{cc}
    \hspace*{-5mm} \includegraphics[width=0.5\linewidth]{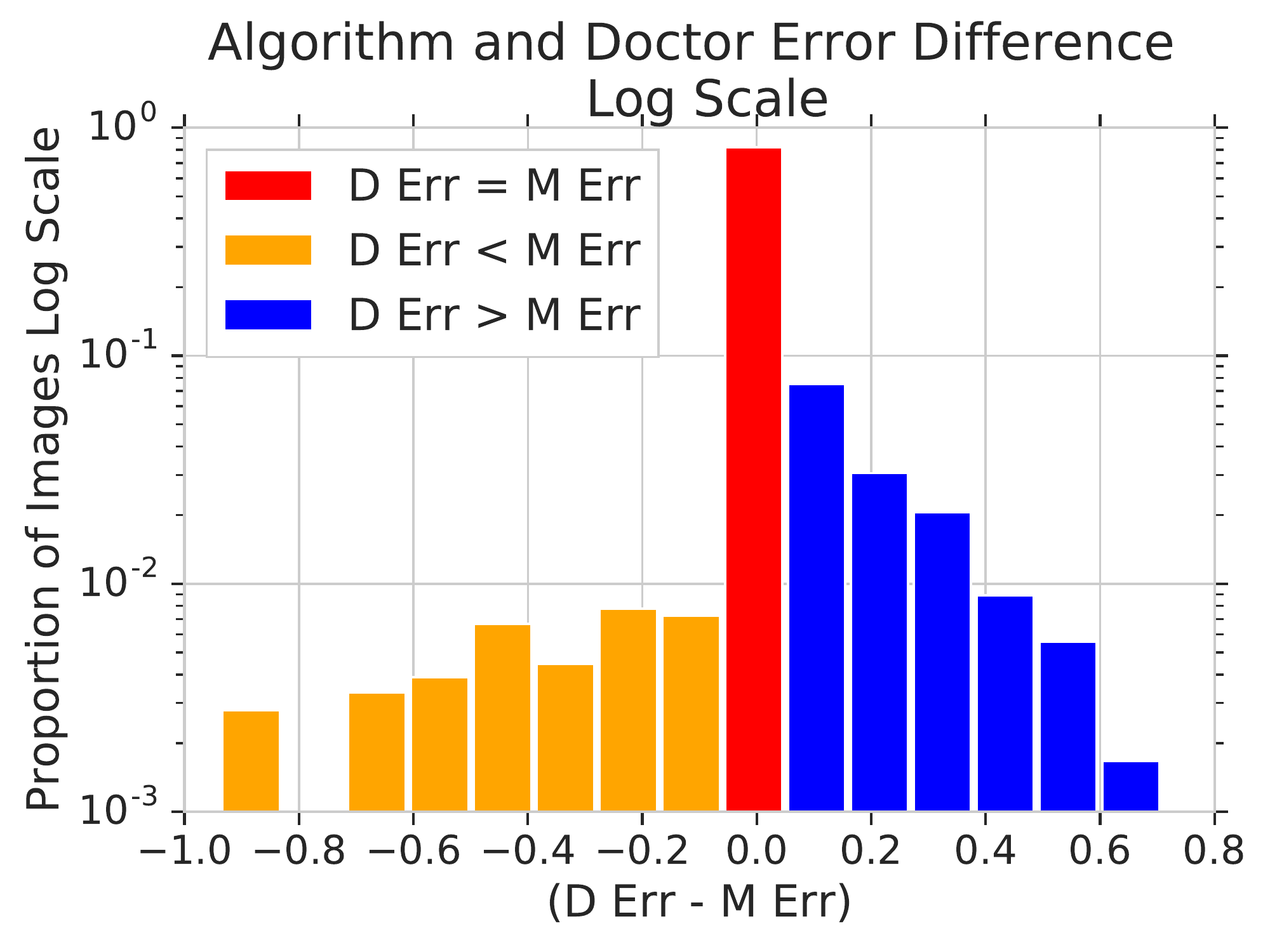} 
    &
    \hspace*{-3mm} \includegraphics[width=0.5\linewidth]{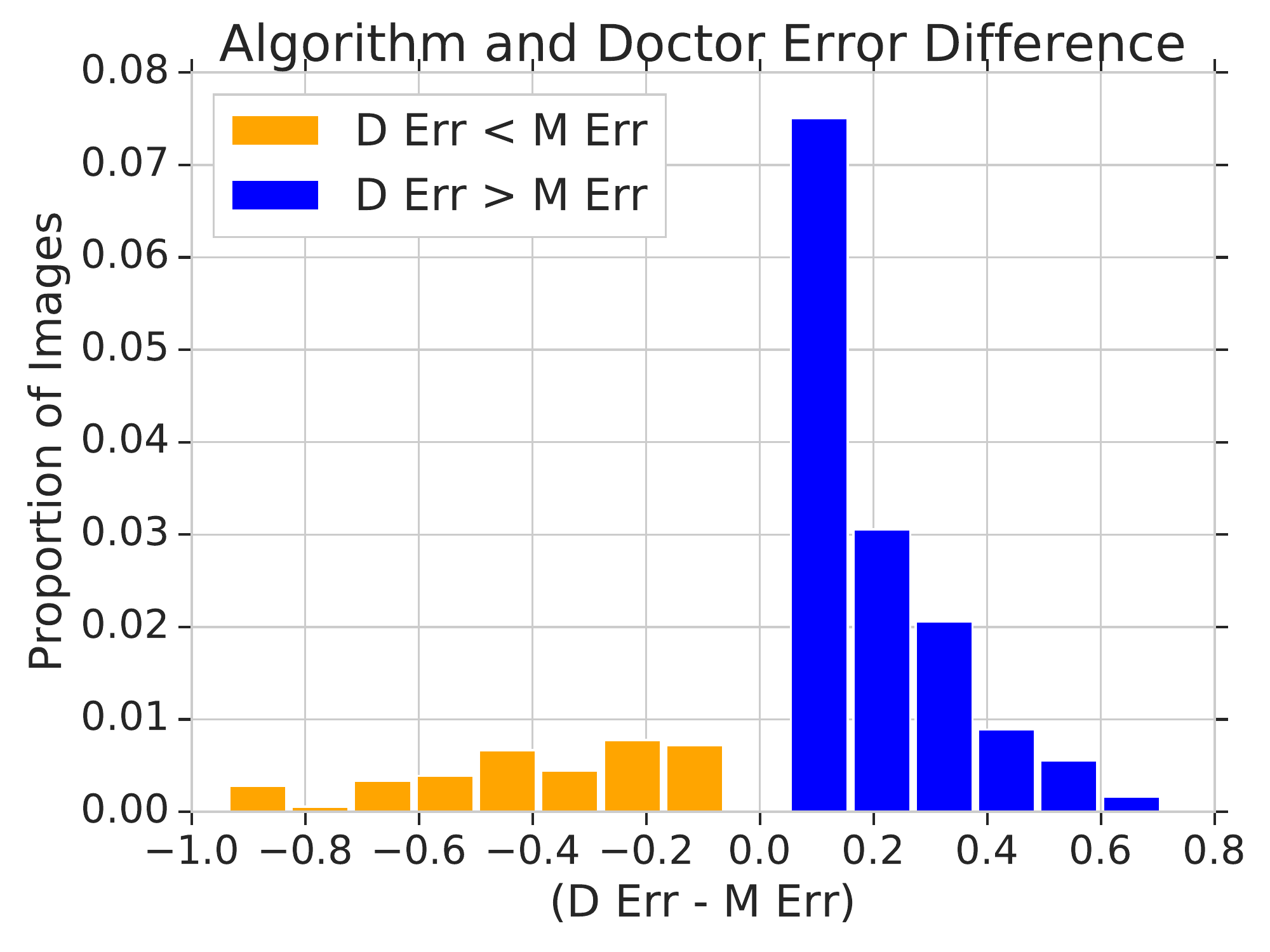} \\
    \end{tabular}
 \caption{\small \textbf{Histogram plot of $\Prb{H_i} - \Prb{M_i}$ for instances $i$ in the adjudicated evaluation dataset.} We show a histogram of probability of human doctor error minus probability of model error over the examples in the adjudicated dataset. The orange bars correspond to examples where the human expert has a lower probability of error than the algorithm, the red where the probability of error is approximately equal, and the blue where the algorithm's probability of error is lower than the human expert's. The left pane is a log plot, and the right is standard scaling (pictured without the red bar.) While the algorithm clearly has lower error probability than the human in more cases, there is a nontrivial mass (~$5\%$) where the human experts have lower error probability than the model.} 
 \label{fig-scatterplot-adj}
\end{figure}

In order to look at the differences in performance between humans
and algorithms on an instance-by-instance level, 
we want to define, for each instance $x_i$ in the adjudicated dataset,
an error probability $\Prb{H_i}$ for the doctors (human experts) and an error probability $\Prb{M_i}$ for 
the algorithm.  

The quantity $\Prb{H_i}$ is straightforward to compute on 
the adjudicated dataset: for an instance $x_i$, suppose that
$n_i$ doctors evaluate it, assigning it 
binary
non-referable/referable grades $h^{(1)}(x_i),...,h^{(n_i)}(x_i)$.
Let $a(x_i)$ be the 
binary adjudicated grade for $x_i$. Then we can define \[ \Prb{H_i} =
\frac{\sum_{j=1}^{n_i} |h^{(j)}(x_i) - a(x_i)|}{n_i}. \] That is,
$\Prb{H_i}$ is the average disagreement of doctors with the
adjudicated grade.

Computing $\Prb{M_i}$ is a little more complicated. Recall that for an
instance $i$, the convolutional neural network model in the algorithm
outputs a value $m(x_i)$ between $[0, 1]$ that is then thresholded to give a
binary decision. A naive estimate of the error probability could
therefore be $m(x_i)$ if the instance is \textit{not} referable, 
and $1 - m(x_i)$ if the instance \textit{is} referable.  Unfortunately, deep
neural networks are well-known to be poorly calibrated
\cite{guo2017calibration}, and this naive approximation is both poorly
calibrated and at a different scaling to the human doctors.
This is not a concern for the algorithm's binary decision, since only the rank-ordering
of the $m(x_i)$ values matter for this, but it poses a challenge
for producing a probability that can serve as $\Prb{M_i}$.

\subsubsection{Determining Algorithm Error Probabilities}
\label{sec-alg-error-probs}
To overcome this issue, we develop a simple method to calibrate the convolutional neural network's output. Recall that the neural network outputs a value $m(x_i) \in [0, 1]$ for each image $x_i$ -- i.e. it induces a ranking over the images $x_i$, which is used to determine the algorithmic decision. We evaluate this induced ranking directly by asking: 

\textit{Suppose we produced a (random) number $R$ of referable instances by sampling a random doctor for each instance, what is the probability that $x_i$ is among the top $R$ instances in the induced ranking?}

We define $\tilde{p}(x_i)$ as the probability that the prediction algorithm declares $x_i$ to be referable. We can then define the error probability, $\Prb{M_i}$, as $\tilde{p}(x_i)$ if the adjudicated grade $a(x_i)$ is referable, and $1 - \tilde{p}(x_i)$ if $a(x_i)$ is non-referable. In Appendix Section \ref{app-sec-model-error}, we provide specific details of the implementation.

\subsubsection{Results on Performance Diversity}
We can now use the estimate of $\Prb{M_i}$ and $\Prb{H_i}$ to study the variation in human expert and algorithmic error across different instances. Specifically, we plot a histogram of values of $\Prb{H_i} - \Prb{M_i}$ across all the adjudicated image instances. 

The result is shown in Figure \ref{fig-scatterplot-adj}. We see that
while there are more images where $\Prb{M_i} < \Prb{H_i}$, there is a
non-trivial fraction of images with $\Prb{M_i} > \Prb{H_i}$. In the
subsequent sections, we analyze different ways of \textit{predicting}
these differences as a way to perform triage, and demonstrate the
resulting gains.

\subsection{Performing Triage and Reallocating Human Effort}
\label{sec-triage-reallocation}

In formulating the basic problem of automation, we considered two
baselines for performance.
The first is {\em full automation}, in which the overall loss
is $\sum_{x_i \in U} g(x_i)$.
The second is equal coverage of all instances by human effort:
if we have a budget of $B = cN$ units of effort for $N$ instances, then
we allocate $c$ units of human effort to each, resulting in a loss
of $\sum_{x_i \in U} f(x_i,c)$.
Our goal here is to show that by allocating human and
algorithmic effort more effectively according to
optimization problem (\ref{eq:opt}) from Section \ref{sec:framework}, 
we can improve on both of these baselines.

Recall the basic heuristic from Section \ref{sec:framework}:
for an arbitrary $\alpha \in [0,1]$, we compute a {\em triage score}
$\tau_\alpha(x_i)$ for each instance $x_i$; we assign the 
first $\alpha N$ to the set $S$ to be handled by the algorithm,
and we allocate equal amounts of human effort
to the remaining set $T$ of $(1 - \alpha) N$ instances.
Note that $\alpha = 1$ corresponds to the full automation baseline,
while $\alpha = 0$ corresponds to equal coverage of all instances 
by human effort.
We will see, however, that stronger performance can be achieved
for intermediate values of $\alpha$.


We begin with two ways of computing the triage score.
The first follows the basic strategy from Section \ref{sec:framework},
where we train two algorithmic predictors to estimate (i) the algorithm's error probability, $\Prb{M_i}$ and (ii) the human error probability $\Prb{H_i}$. Specifically, we train two auxillary neural networks, one to predict $\Prb{H_i}$ and one to predict $\Prb{M_i}$. To predict $\Prb{H_i}$, we build off of the work of \cite{raghu2018DUP} on direct prediction of doctor disagreement: we label each example with a $0$ if there is agreement amongst the doctor grades, and $1$ otherwise, and train a small neural network to predict these agreement labels from the image embedding. A similar setup is employed for predicting $\Prb{M_i}$, where the binary label now corresponds to whether the output of the diagnostic 5-class convolutional neural network agrees with the doctor grades -- i.e. does the model make an error on that image. The full details of this process are described in Appendix \ref{app-sec-model-error}. 

The second method of computing a triage score establishes an ``ideal'' benchmark on the potential power of the optimization problem (\ref{eq:opt}) using aspects
of ground truth, sorting the instances by the true value of $\Prb{H_i} - \Prb{M_i}$, since this divides the instances between humans and algorithms based on the relative strength of each party on the respective instances.

\begin{figure}
  \centering
  \begin{tabular}{cc}
    \hspace*{-5mm} \includegraphics[width=0.5\linewidth]{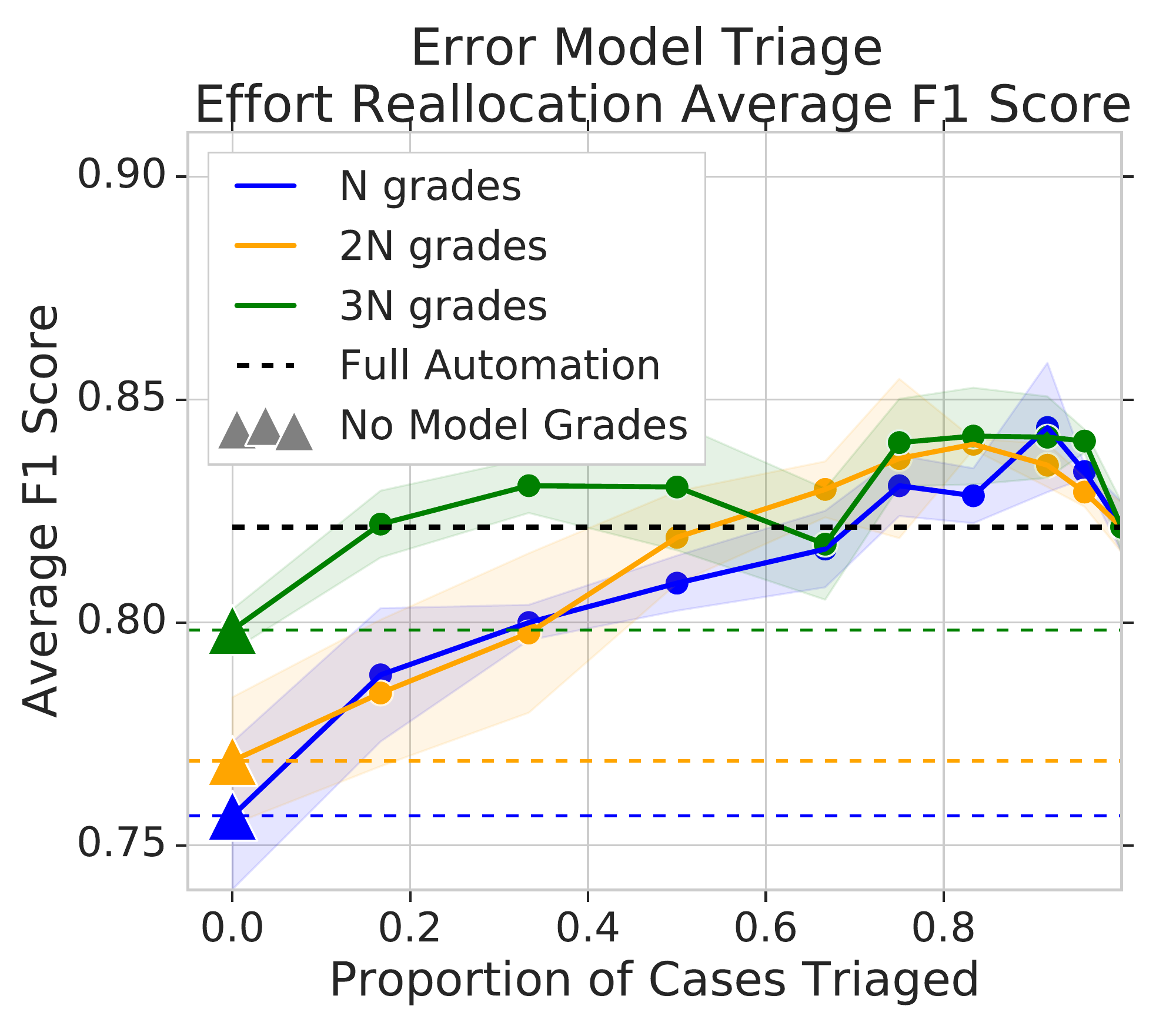} 
    &
    \hspace*{-3mm} \includegraphics[width=0.5\linewidth]{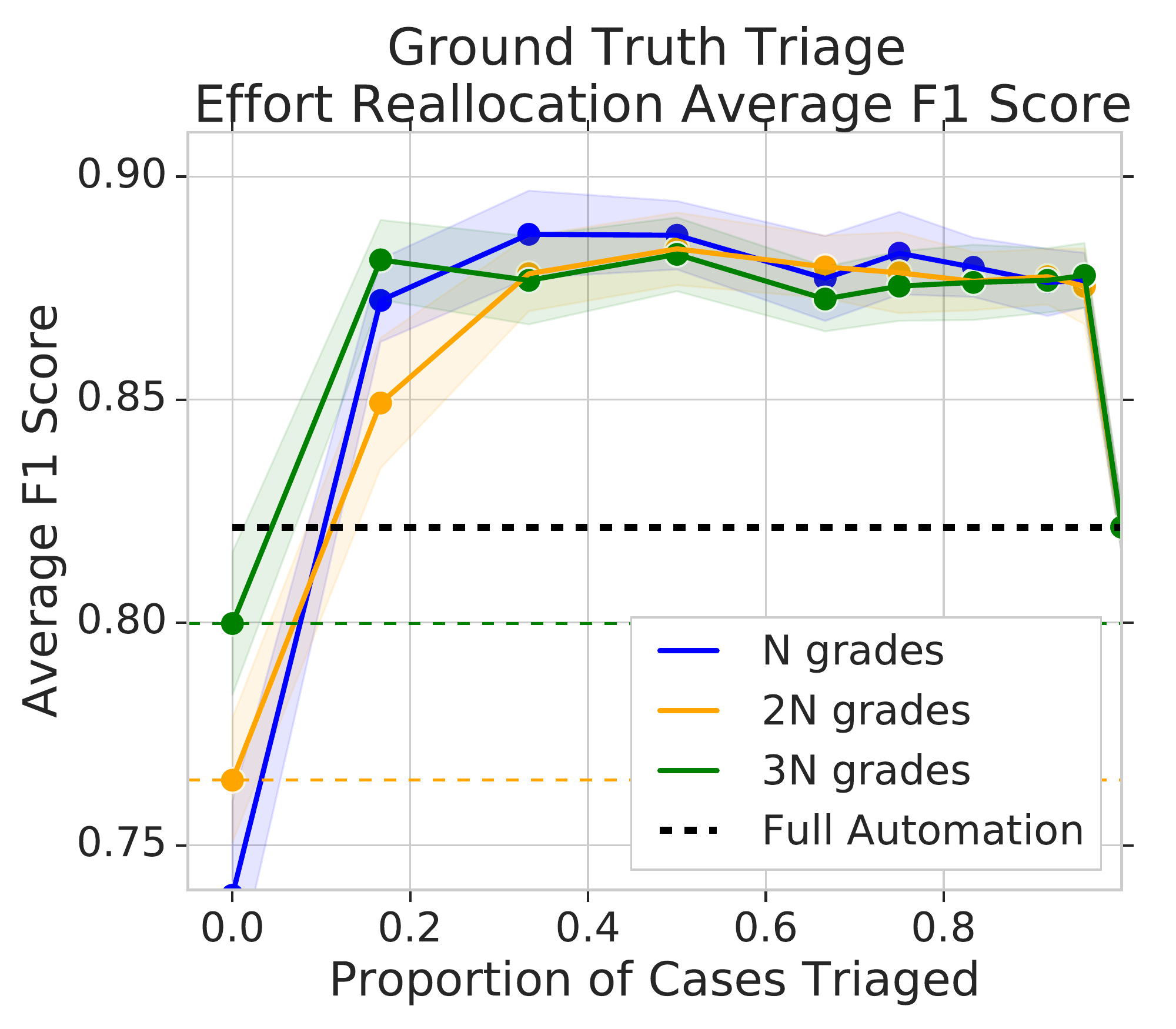} \\
    \hspace*{-5mm} \includegraphics[width=0.5\linewidth]{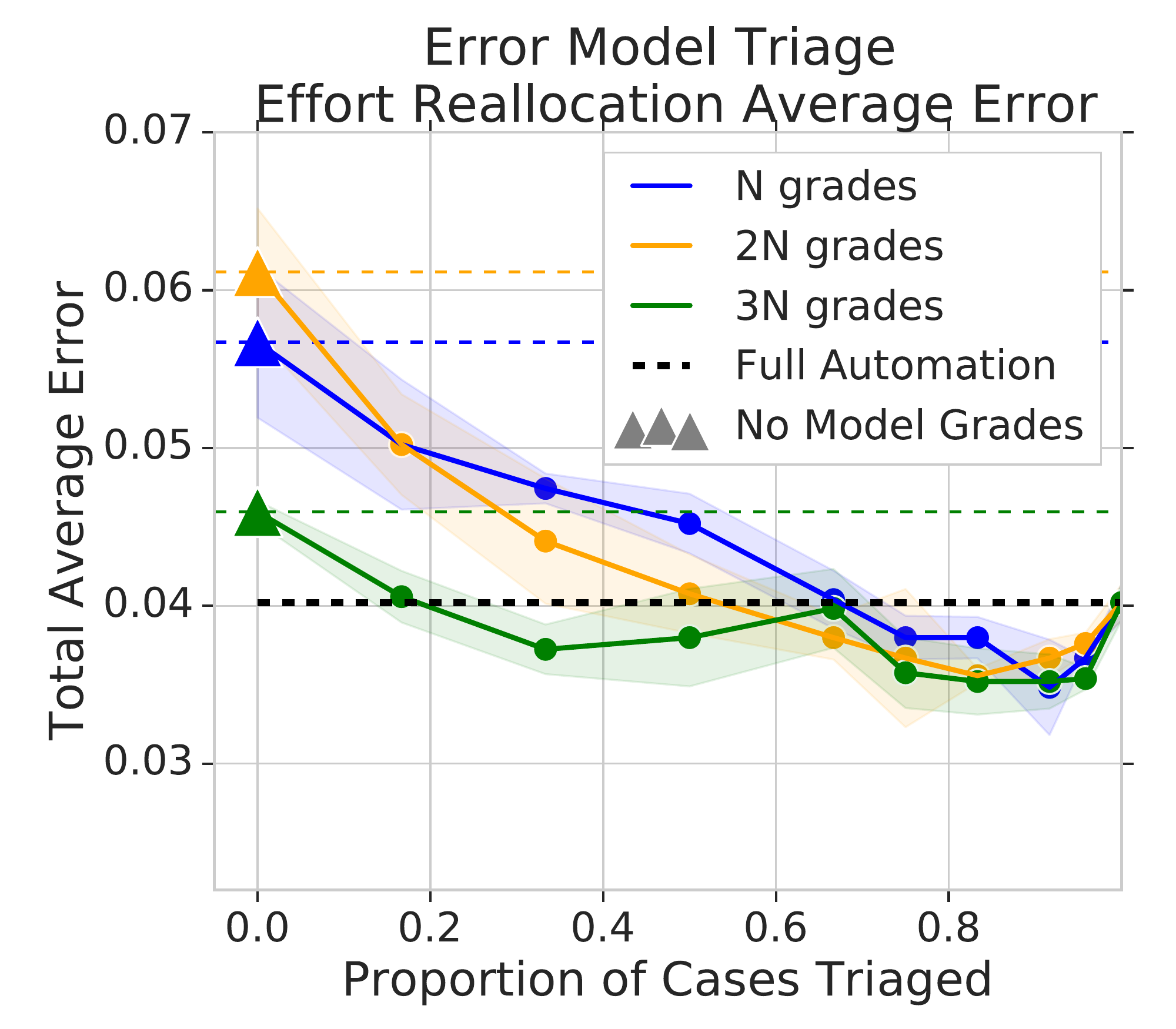} 
    &
    \hspace*{-3mm} \includegraphics[width=0.5\linewidth]{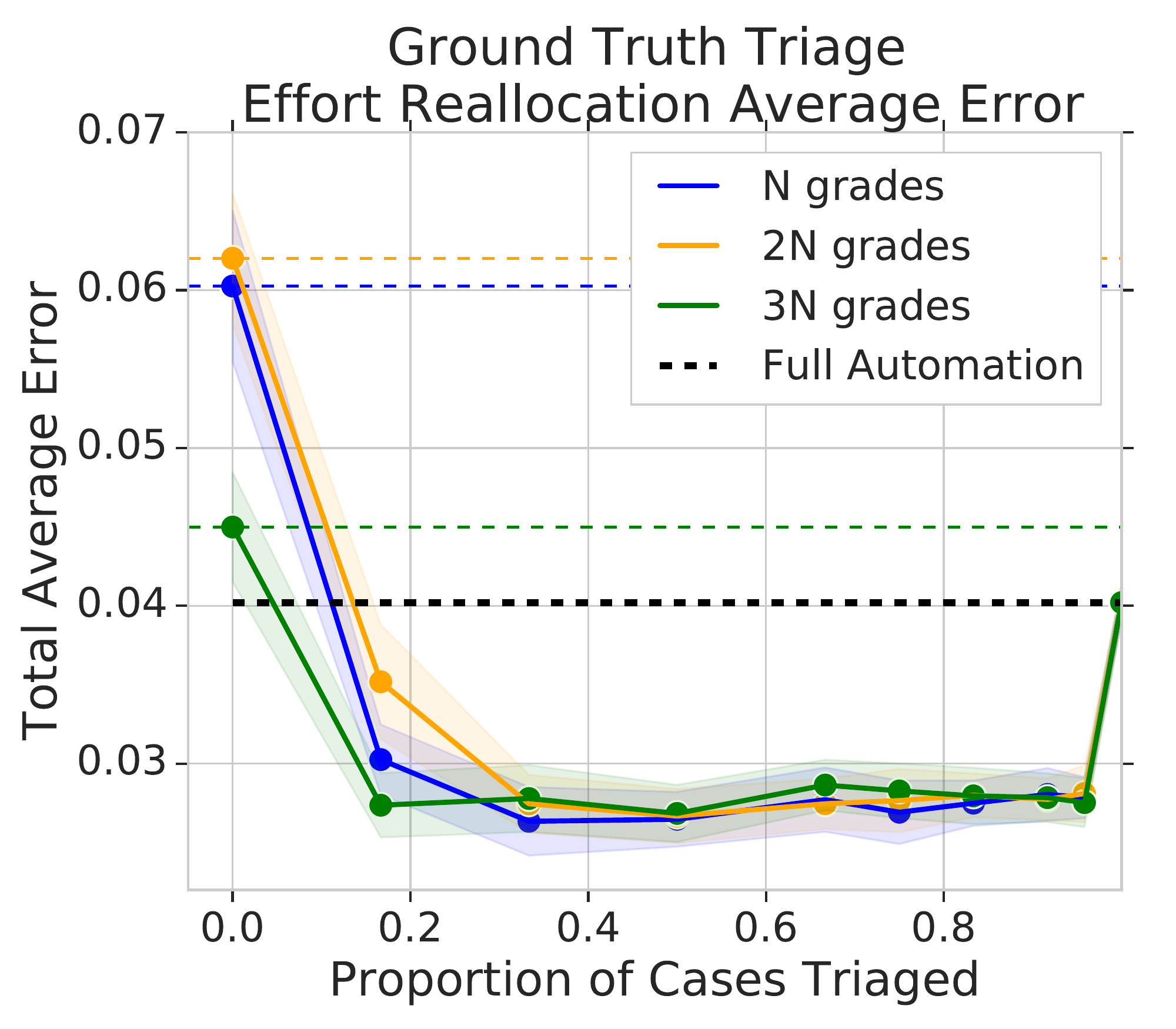}
    \end{tabular}
 \caption{\small \textbf{Combing algorithmic and human effort by
 triaging outperforms full automation and the equal coverage human
 baseline. Left column: triage by the difference between the predicted values of $\Prb{H_i}$ and $\Prb{M_i}$. Right column:
 triage by ground truth $\Prb{H_i} - \Prb{M_i}$} We order the images
 by their triage scores (predicted $\Prb{H_i} - \Prb{M_i}$ for the left column and ground truth
 $\Prb{H_i} - \Prb{M_i}$ on the right), and automate an $\alpha$
 fraction of them. The remaining $(1 - \alpha)N$ images have the human
 doctor budget ($N, 2N, 3N$ grades) allocated amongst them, according to the equal coverage protocol. This is described in further detail in Appendix Section \ref{app-sec-alloc-alg}. The black dotted line is the performance of full automation, and the coloured
 dotted lines the performance of equal coverage for the different
 total number of doctor grades available. We see
 that triaging and combining algorithmic and human effort performs
 better than all of these baselines. Triaging by
 ground truth (right column) gives significant gains, and suggests
 that better triage prediction is a crucial problem that merits
 further study. In Appendix Section \ref{app-sec-alg-grades}, we also include results when the remaining $(1 - \alpha)N$ cases have the algorithm grade available, along with the reallocated human effort. The qualitative conclusions are identical.}
 \label{fig-triage-effort-reallocation}
\end{figure}

In both cases, we determine the performance of the human effort
using the average of a corresponding number of
randomly sampled doctor grades from the data.
This allows us to demonstrate improvements without
any assumptions on how the doctors might use information from the
algorithmic predictions on these instances.
It is also reasonable, however, to imagine a scenario in 
which the algorithmic predictions are still freely available
even on the instances that we assign to the human doctors, and
to consider simple models for how the doctor grades might be combined 
with these algorithmic predictions.
We consider this case in the Appendix, which supports the same conclusions.

\subsubsection{Triage Results}
The results for these two triage scores, as we vary $\alpha$,
are shown in Figure \ref{fig-triage-effort-reallocation}. 
The figure depicts both the 
average error (bottom row), as well as the F1 score (top row), which
accounts for imbalances between the number of referable and
non-referable instances.
The left column corresponds to using the difference between the predicted values of $\Prb{H_i}$ and $\Prb{M_i}$ as a triage score, while the right column corresponds to using the true value $\Prb{H_i} - \Prb{M_i}$ to perform triage. In both triage
schemes, we observe that the best performance comes for $0 < \alpha <
1$, beating both the full automation protocol (dotted black line) and
equal coverage of all instances by human effort (coloured dotted lines). 

While combining algorithmic and human effort in both of these ways leads to performance gains, we see that the ground
truth triage ordering performs significantly better than triaging by the predicted error probability. This suggests that learning better triage predictors
might have an even greater impact on overall deployed performance than
continuous slight improvements to diagnostic accuracy.

\subsubsection{The Simplest Heuristic: Algorithmic Uncertainty}
\label{sec-triage-model-effort-realloc}

\begin{figure}
  \centering
  \begin{tabular}{cc}
\hspace*{-5mm} \includegraphics[width=0.5\linewidth]{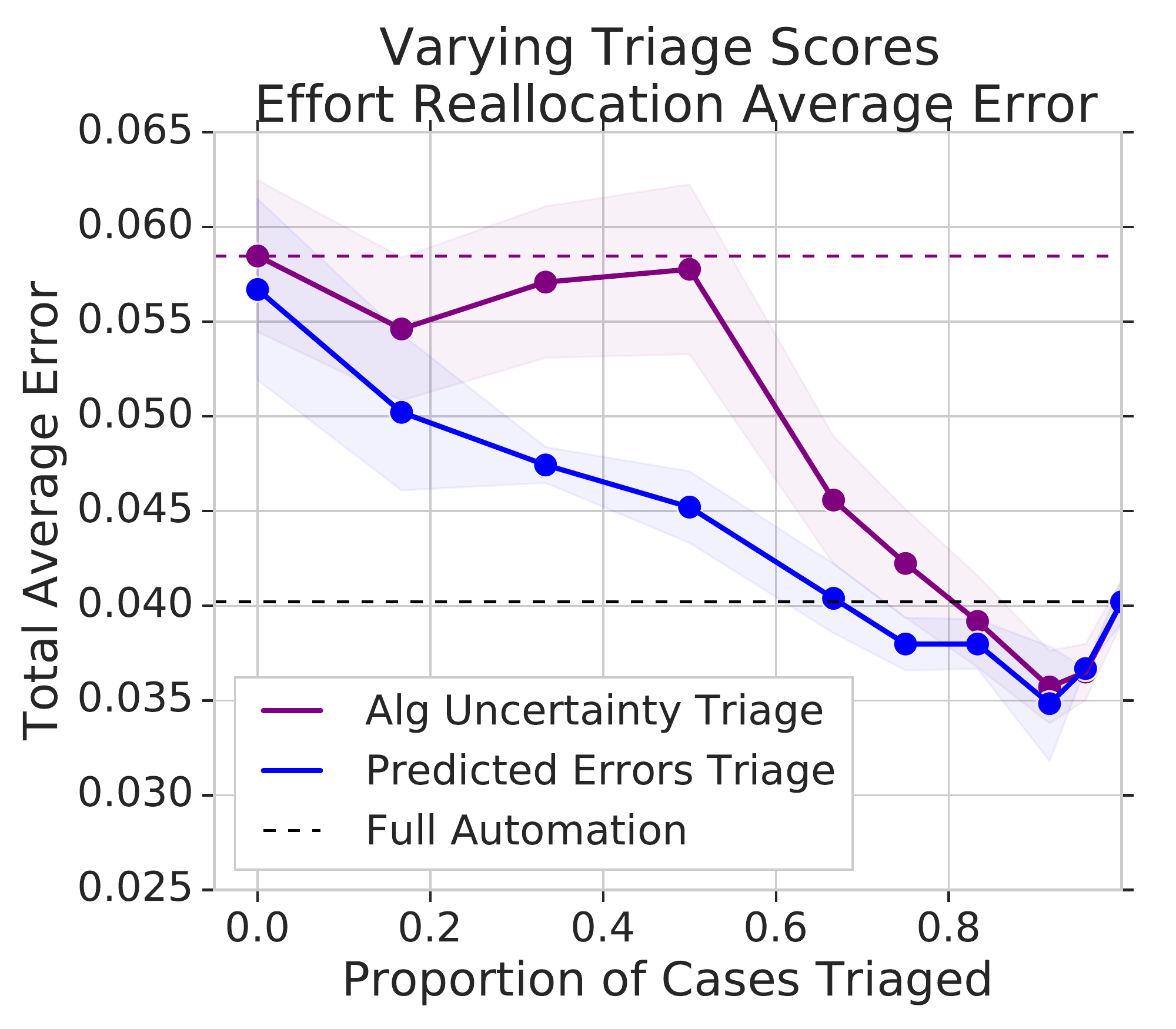} &
\hspace*{-3mm} \includegraphics[width=0.5\linewidth]{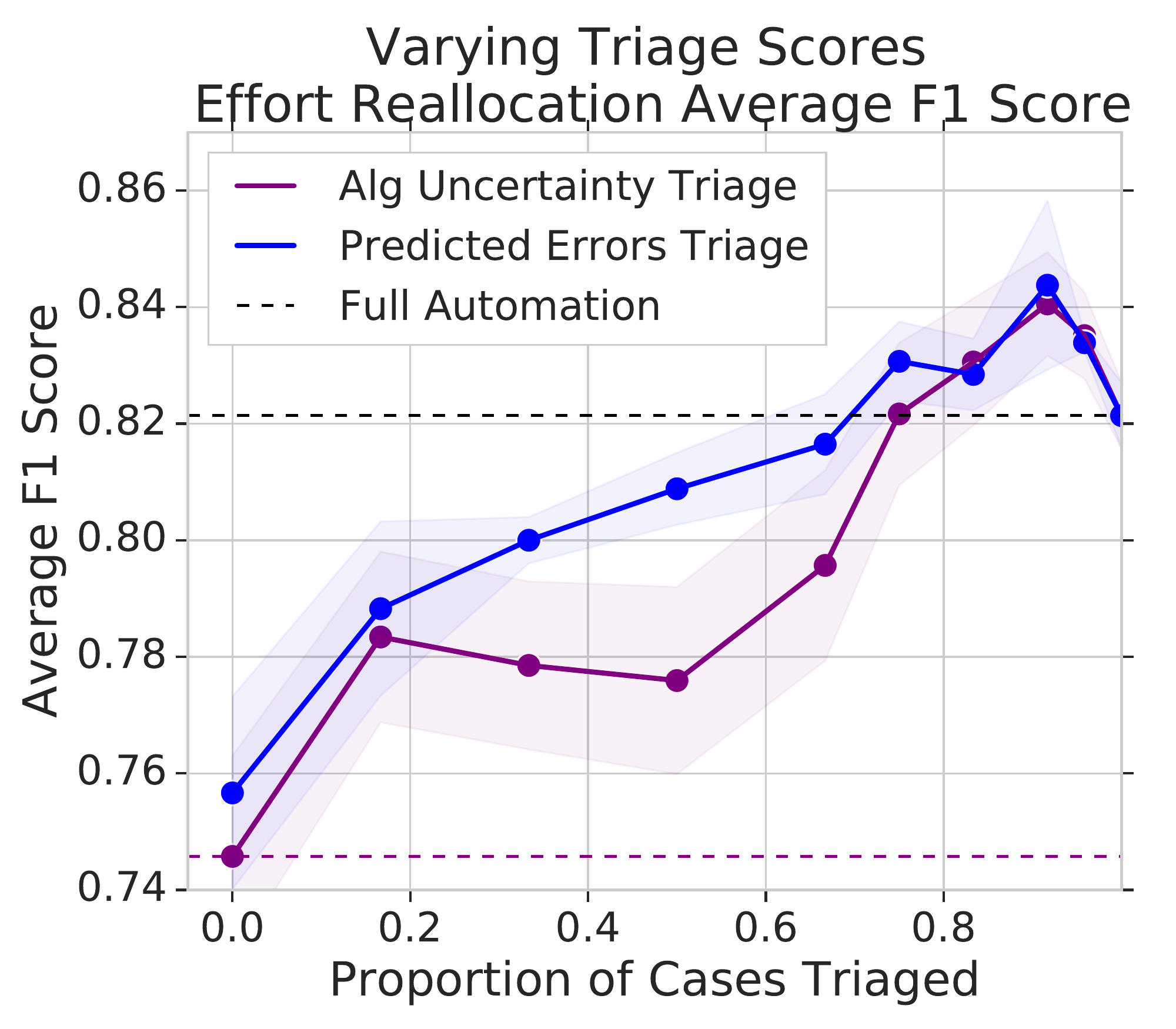} 
 \end{tabular}
 \caption{\small \textbf{Even triaging by algorithm uncertainty leads to gains over pure algorithmic and pure human performance.} Instead of the separate error prediction algorithms, we triage by the simple algorithm uncertainty: $m(x)(1 - m(x))$, which acts as a proxy for algorithm error probability (and no explicit modelling of human error probability.) The same qualitative conclusions hold with this simple triage score also (purple line), although larger gains are achieved with the separate error prediction algorithms (blue line). These results are for $N$ doctor grades, the same conclusions hold for $2N, 3N$ grades.} 
 \label{fig-triage-model-effort-realloc}
\end{figure}

In the previous section, we saw the results of training two separate algorithmic predictors to estimate the values of $\Prb{H_i}$ and $\Prb{M_i}$, and using the difference between these predicted values as a triage score. An even simpler triage score is given by only using the algorithm's uncertainty, $m(x)(1 - m(x))$. In Figure \ref{fig-triage-model-effort-realloc}, we show that even this triage score, available `for free' from the algorithmic predictor, improves upon pure automation and pure human effort, although larger gains are available through using the two algorithmic error predictors. These results reiterate the rich possibilities for gains from algorithmic triage.

\subsection{Differential Costs and Zero-Error Subsets}
\label{sec-automation}

Finally, we recall a further consideration from the 
framework in Section \ref{sec:framework}:
suppose the social planner views errors made by algorithms as
more costly than errors made by humans, 
resulting in an objective function 
of the form in (\ref{eq:opt-lambda}),
$\lambda \sum_{x \in S} g(x) + \sum_{x \in T} f(x,k_x)$.
As $\lambda$ becomes large, what does this imply about
the use of algorithmic predictions?

We find in our application that it is possible to identify 
large subsets of the data on which the algorithm achieves
{\em zero error}.
Such a fact can easily be hidden by considering only
aggregate measures of algorithmic performance, and it implies 
that even when $\lambda$ is large, there may still be an 
important role for algorithms in automation.

To quantify this effect, we order the instances by a triage
score as in our earlier analyses.
We then look at the average error of the algorithmic
predictions on the first $\alpha$ fraction of images:
for $\alpha$ varying between $0$ and $1$, we plot
\[ \frac{M_{err}(\alpha N)}{N} \]
where $M_{err}(\alpha N)$ is the number of errors 
made by the model on the first $\alpha N$ instances.

\begin{figure}
  \centering
  \begin{tabular}{cc}
\hspace*{-5mm} \includegraphics[width=0.5\linewidth]{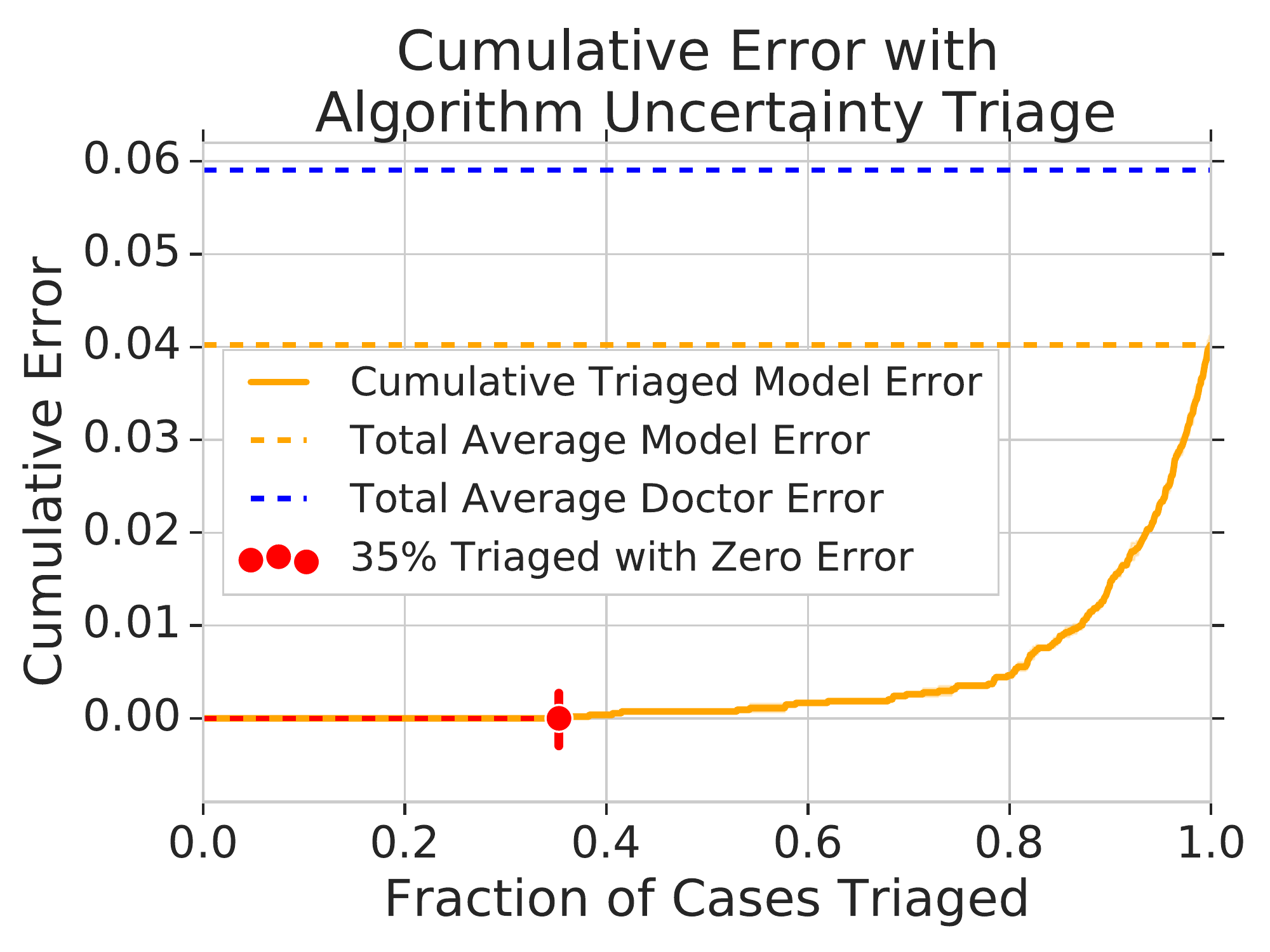} &
\hspace*{-3mm} \includegraphics[width=0.5\linewidth]{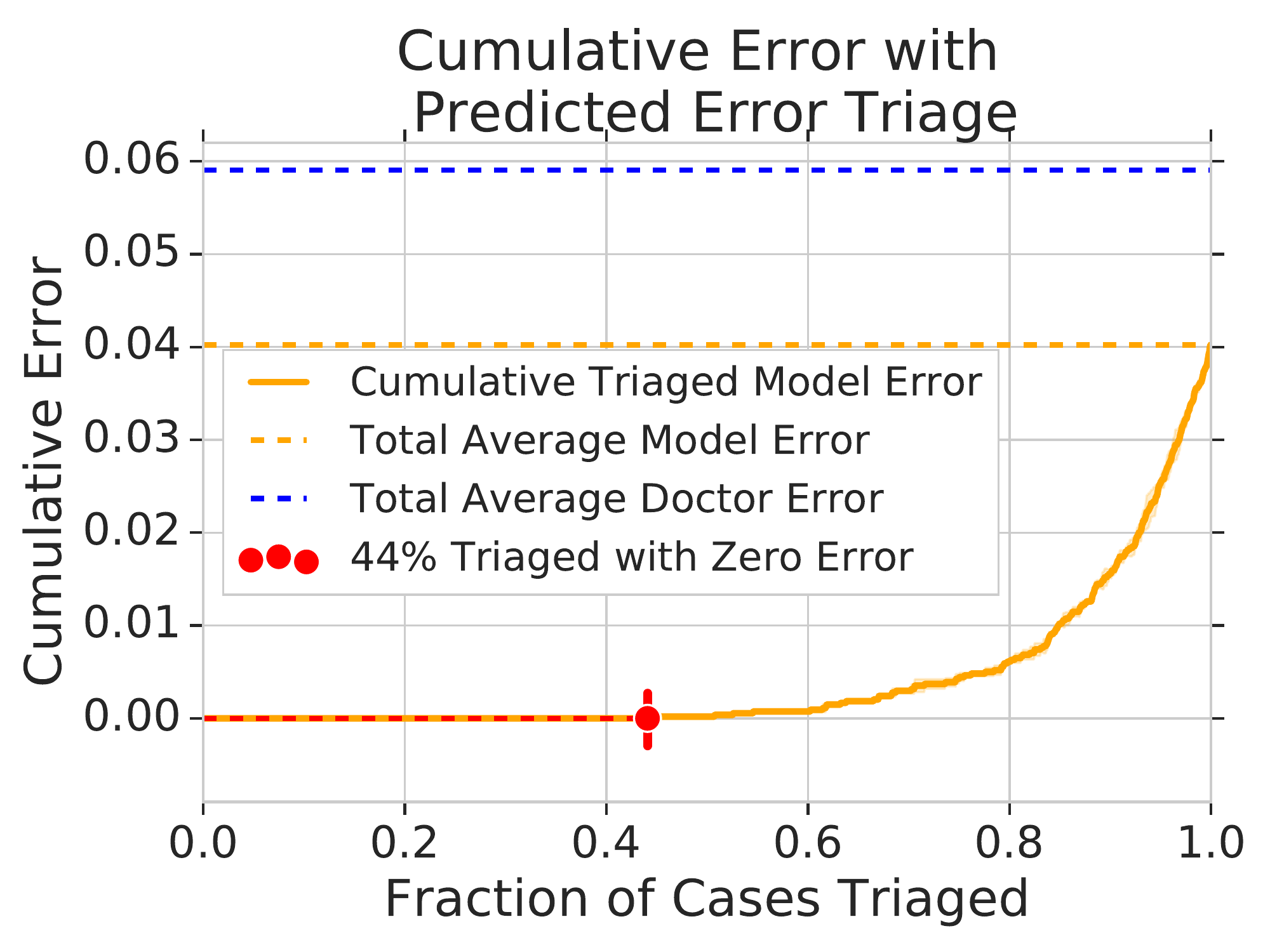} 
 \end{tabular}
 \caption{\small \textbf{Triage identifies large subsets of the data
 with zero error.} We plot the average cumulative error
 $\frac{M_{err}(\alpha N)}{N}$, where $M_{err}(\alpha N)$ is the
 number of errors made by the algorithm on the first $\alpha$ fraction
 of the $N$ images when triaged. We observe that triaging even by the
 simple uncertainty measure, $m(x_i)(1 - m(x_i))$ (left plot), can identify
 a $35\%$ fraction of data where the algorithm makes zero errors.
 Using the separate error prediction model from Section
 \ref{sec-triage-model-effort-realloc}, we can improve on this, identifying
 $44\%$ of the data where the algorithm has zero errors. The plot is
 averaged over three repetitions (so each repeat identified at least
 $35\%, 44\%$ of the data respectively.)} \label{fig-effort-saving}
\end{figure}

\subsubsection{Results}
\label{sec-model-err-prediction}
Figure \ref{fig-effort-saving} left pane shows the results of plotting this quantity. We triage the cases both by our prediction of $\Prb{H_i} - \Prb{M_i}$ from the two error prediction algorithms as well as the simple algorithm uncertainty term, $m(x)(1 - m(x))$. We evaluate the average error of the algorithmic predictions on the first $\alpha$ fraction of images, over three repetitions of training the diagnostic neural network component of the algorithm. We see that even using the simple $m9x)(1 - m(x))$ as a triage score, we can identify a zero-error subset that is $35\%$ the size of the entire dataset. Similar to Section \ref{sec-triage-model-effort-realloc}, further improvements are shown by predicting the value of $\Prb{H_i} - \Prb{M_i}$. The right pane of Figure \ref{fig-effort-saving} shows this result, where we can identify a zero-error subset of size $44\%$, again averaged over three repetitions. 

\section{Related Work}
With the successes of machine learning and particularly deep learning methodologies in modalities such as imaging, there have been numerous works comparing algorithmic performance to human performance in medical tasks, albeit in frameworks that implicitly interpret automation as success in prediction. In this prediction setting, the general comparison is between the case in which only the algorithm is used and the case in which only human effort is used; such comparisons have been done for chest x-rays \cite{rajpurkar2017chexnet}, for Alzheimer's detection from PET scans \cite{ding2018deep}, and for the setting we consider here based on diabetic retinopathy diagnosis from fundus photographs (and OCT scans) \cite{de2018clinically,Gulshan2016Retinal}.  The recent survey paper by Topol \cite{topol2019medicine} references several additional studies of this kind. A few papers have begun to look at fixed modes of interaction with humans, including processes in which algorithmic outputs are reviewed by physicians \cite{cai2019human,deniz2018segmentation,liu2017detecting}, as well as fixed combinations of physician and algorithmic judgments \cite{raghu2018model}.

\section{Discussion}
This work has presented a framework for analyzing automation by algorithms.
Rather than treating the introduction of algorithms in an all-or-nothing
fashion, we show that stronger performance can be obtained if algorithms
are used both (i) for prediction on instances of the problem, and (ii) for
providing triage judgments about which instances should be handled
algorithmically and which should be handled by human effort.
This broader formulation of the automation question highlights the
importance of accurately estimating the propensity of both humans
and algorithms to make errors on a per-instance basis, and the
use of these estimates in an optimization framework for allocating
effort efficiently.
Analysis of an application in diabetic retinopathy diagnosis 
shows that this framework can lead to performance gains even
for well-studied problems in AI applications in medicine.

Through the analysis of benchmarks for stronger performance, we
also highlight how stronger predictions of per-instance error
has the potential to yield still better performance.
Our findings thus demonstrate how further study of algorithmic
triage and its role in allocating human and computational effort
has the potential to yield substantial benefits for the task of automation.

\section*{Acknowledgements}
We thank Vincent Vanhoucke, Quoc Le, Yun Liu and Samy Bengio for helpful feedback.  

\bibliographystyle{plain}
\bibliography{references}

\begin{thebibliography}{10}

\bibitem{ahsan2015dr}
Hasseb Ahsan.
\newblock Diabetic retinopathy -- biomolecules and multiple pathophysiology.
\newblock {\em Diabetes and Metabolic Syndrome: Clincal Research and Review},
  pages 51--54, 2015.

\bibitem{ICDRStandards}
American Academy of Ophthalmology.
\newblock {\em International Clinical Diabetic Retinopathy Disease Severity
  Scale Detailed Table}.

\bibitem{cai2019human}
Carrie~J Cai, Emily Reif, Narayan Hegde, Jason Hipp, Been Kim, Daniel Smilkov,
  Martin Wattenberg, Fernanda Viegas, Greg~S Corrado, Martin~C Stumpe, et~al.
\newblock Human-centered tools for coping with imperfect algorithms during
  medical decision-making.
\newblock {\em arXiv preprint arXiv:1902.02960}, 2019.

\bibitem{de2018clinically}
Jeffrey De~Fauw, Joseph~R Ledsam, Bernardino Romera-Paredes, Stanislav Nikolov,
  Nenad Tomasev, Sam Blackwell, Harry Askham, Xavier Glorot, Brendan
  O’Donoghue, Daniel Visentin, et~al.
\newblock Clinically applicable deep learning for diagnosis and referral in
  retinal disease.
\newblock {\em Nature medicine}, 24(9):1342, 2018.

\bibitem{deniz2018segmentation}
Cem~M Deniz, Siyuan Xiang, R~Spencer Hallyburton, Arakua Welbeck, James~S Babb,
  Stephen Honig, Kyunghyun Cho, and Gregory Chang.
\newblock Segmentation of the proximal femur from mr images using deep
  convolutional neural networks.
\newblock {\em Scientific reports}, 8(1):16485, 2018.

\bibitem{ding2018deep}
Yiming Ding, Jae~Ho Sohn, Michael~G Kawczynski, Hari Trivedi, Roy Harnish,
  Nathaniel~W Jenkins, Dmytro Lituiev, Timothy~P Copeland, Mariam~S Aboian,
  Carina Mari~Aparici, et~al.
\newblock A deep learning model to predict a diagnosis of alzheimer disease by
  using 18f-fdg pet of the brain.
\newblock {\em Radiology}, 290(2):456--464, 2018.

\bibitem{Gulshan2016Retinal}
Varun Gulshan, Lily Peng, Marc Coram, Martin~C Stumpe, Derek Wu, Arunachalam
  Narayanaswamy, Subhashini Venugopalan, Kasumi Widner, Tom Madams, Jorge
  Cuadros, Ramasamy Kim, Rajiv Raman, Philip~Q Nelson, Jessica Mega, and Dale
  Webster.
\newblock Development and validation of a deep learning algorithm for detection
  of diabetic retinopathy in retinal fundus photographs.
\newblock {\em JAMA}, 316(22):2402--2410, 2016.

\bibitem{guo2017calibration}
Chuan Guo, Geoff Pleiss, Yu~Sun, and Kilian~Q. Weinberger.
\newblock On calibration of modern neural networks.
\newblock abs/1706.04599, 2017.

\bibitem{kingma2014adam}
Diederik~P Kingma and Jimmy Ba.
\newblock Adam: A method for stochastic optimization.
\newblock {\em arXiv preprint arXiv:1412.6980}, 2014.

\bibitem{krause2018adj}
Jonathan Krause, Varun Gulshan, Ehsan Rahimy, Peter Karth, Kasumi Widner,
  Gregory~S. Corrado, Lily Peng, and Dale~R. Webster.
\newblock Grader variability and the importance of reference standards for
  evaluating machine learning models for diabetic retinopathy.
\newblock abs/1710.01711, 2017.

\bibitem{liu2017detecting}
Yun Liu, Krishna Gadepalli, Mohammad Norouzi, George~E Dahl, Timo Kohlberger,
  Aleksey Boyko, Subhashini Venugopalan, Aleksei Timofeev, Philip~Q Nelson,
  Greg~S Corrado, et~al.
\newblock Detecting cancer metastases on gigapixel pathology images.
\newblock {\em arXiv preprint arXiv:1703.02442}, 2017.

\bibitem{raghu2018model}
Aniruddh Raghu, Matthieu Komorowski, and Sumeetpal Singh.
\newblock Model-based reinforcement learning for sepsis treatment.
\newblock {\em arXiv preprint arXiv:1811.09602}, 2018.

\bibitem{raghu2018DUP}
Maithra Raghu, Katy Blumer, Rory Sayres, Ziad Obermeyer, Sendhil Mullainathan,
  and Jon Kleinberg.
\newblock Direct uncertainty prediction with applications to healthcare.
\newblock {\em arXiv preprint arXiv:1807.01771}, 2018.

\bibitem{rajpurkar2017chexnet}
Pranav Rajpurkar, Jeremy Irvin, Kaylie Zhu, Brandon Yang, Hershel Mehta, Tony
  Duan, Daisy Ding, Aarti Bagul, Curtis Langlotz, Katie Shpanskaya, Matthew~P.
  Lungren, and Andrew~Y. Ng.
\newblock Chexnet: Radiologist-level pneumonia detection on chest x-rays with
  deep learning.
\newblock abs/1711.05225, 2017.

\bibitem{topol2019medicine}
Eric Topol.
\newblock High-performance medicine: the convergence of human and artificial
  intelligence.
\newblock {\em Nature Medicine}, 25:44--56, 2019.

\end{thebibliography}

\clearpage

\appendix

\section{Training Data and Models Details}
\label{app-sec-train-model}
Our training dataset consists of fundus photographs with labels corresponding to individual doctor grades. There are 5 possible DR grades and hence 5 possible class labels. A subset of this data has fundus photographs with more than one doctor grade, corresponding to multiple doctors individually and independently deciding on the grade for the image. The label for these images is not a one-hot class label but the empirical distribution of grades. For example, if an image $i$ has grades $\{2, 3, 3\}$, then its label would be $[0, 1./3, 2./3, 0, 0]$.  

On this data, we train a convolutional neural network, an Inception-v3 model with weights pretrained from ImageNet and a new five class classification head. We train with the Adam optimizer \cite{kingma2014adam} and an initial learning rate of 0.005. To better calibrate the model, we retrain the very top of the network (from the PreLogits layer) on just the data with two or more doctor grades.

\paragraph{Training Error Probability Prediction Models} For Figures \ref{fig-triage-effort-reallocation}, \ref{fig-triage-model-effort-realloc} and \ref{fig-effort-saving}, we use separate error probability prediction algorithms to predict the values of $\Prb{H_i}$ and $\Prb{M_i}$. The setup for predicting $\Prb{M_i}$ is as follows: after training the main convolutional neural network on the train dataset, we train a small fully connected deep neural network to take the prelogit embeddings of a train image $x_i$, and predict whether or not the main convolutional neural network was correct on that image. The label for the image is binary: agree/disagree on whether the mass $m(x_i)$ put on referable by the convolutional neural network thresholded at $0.5$ equals the mass on referable by the human doctor grades, again thresholded at $0.5$.

The setup for training $\Prb{H_i}$ builds off of \cite{raghu2018DUP}. First, we only select cases for which we have at least two doctor grades. For these, we take the image embedding from the Prelogit layer of the large diagnostic convolutional neural network as input, and the label as a binary target. This label is defined as follows: we split the available doctor grades into two evenly sized sets $A$ and $B$. We aggregate all the grades in $A$ into a single referable/non-referable grade by averaging and thresholding at $0.5$, and do the same for the grades in $B$. If these two aggregated grades agree, we label the image with $0$ (agreement, low doctor error probability), if not, we label with $1$ (disagreement, high doctor error probability.)  

\section{Computing $\Prb{M_i}$}
\label{app-sec-model-error}
In Section \ref{sec-alg-error-probs}, we overviewed the method used to define a well calibrated error probability for the output of the convolutional neural network. In Algorithm \label{calibrate-model}, we give a step-by-step overview of the implementation of this method. In our experiments, we set $C = 2000$. 

\begin{algorithm}[t]
  \caption{Model Error Probability Calibration}\label{calibrate-model}
  \begin{algorithmic}[1]
    \State $Err_i = 0$ for $i$ in instances.
    \For{($r=0$; $r<C$; $r$++)}\Comment{$C$ is a sufficiently large constant.}
      \State $R = 0$
      \For{$i$ \texttt{in instances}}\Comment{Sample a doctor grade and count number of referables.}
        \State \texttt{Sample doctor grade} $h^{(i)}$
        \If{($h^{(i)} \geq 3$)}
            \State $R\gets R + 1$
        \EndIf
       \EndFor
       \State \texttt{Rank instances $i$ from highest to lowest $m(x_i)$}
       \State $A_R = \{i : rank(i) \leq R \}$
       \For{$i$ \texttt{in instances}} \Comment{Assign binary grades to instances. Top $R$ referable.}
        \If{($i \in A_R$)}
            \State $m_i\gets 1$
        \Else
            \State $m_i\gets 0$
        \EndIf
        \State $Err_i\gets Err_i + |m_i - a_i|$\Comment{$a_i$ is the adjudicated grade}
        \EndFor
    \EndFor
    \State \textbf{return} $\Prb{M_i} = Err_i/C$\Comment{Error probability by averaging over number of repetitions.}
  \end{algorithmic}
\end{algorithm}

\section{Triage and Allocation Algorithm}
\label{app-sec-alloc-alg}
When using triage to reallocate human effort, we first order the instances by their triage scores, and then fully automate the first $\alpha N$ of them. On the remainder $(1 - \alpha)N$ images, we allocate the budget of $cN$ human doctor grades we have available. To allocate this set of $cN$ grades, we use the equal coverage protocol: each of the remaining $(1 - \alpha)N$ cases gets $cN/((1- \alpha)N)$ grades. If this is a non-integer amount, with $r$ spare grades, the $r$ cases identified as the hardest (according to the triage scores) get an additional grade. We then compute the final binary decision by taking the mean grade (for each case) and thresholding by $0.5$ (the majority vote.)

\subsection{Results on other Thresholds}
\begin{figure}
  \centering
  \begin{tabular}{cccc}
    \hspace*{-10mm} \includegraphics[width=0.25\linewidth]{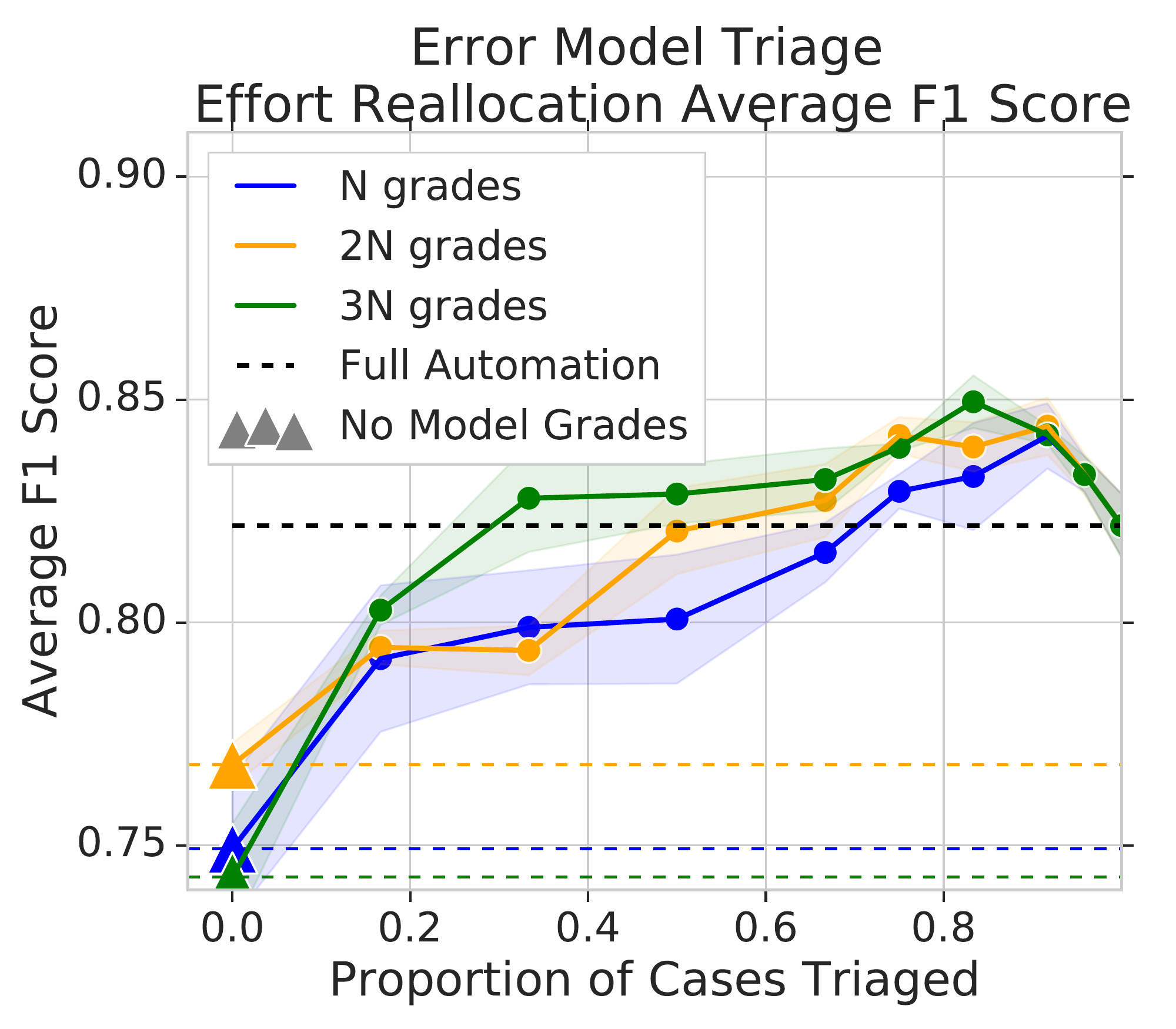} 
    &
    \hspace*{-6mm} \includegraphics[width=0.25\linewidth]{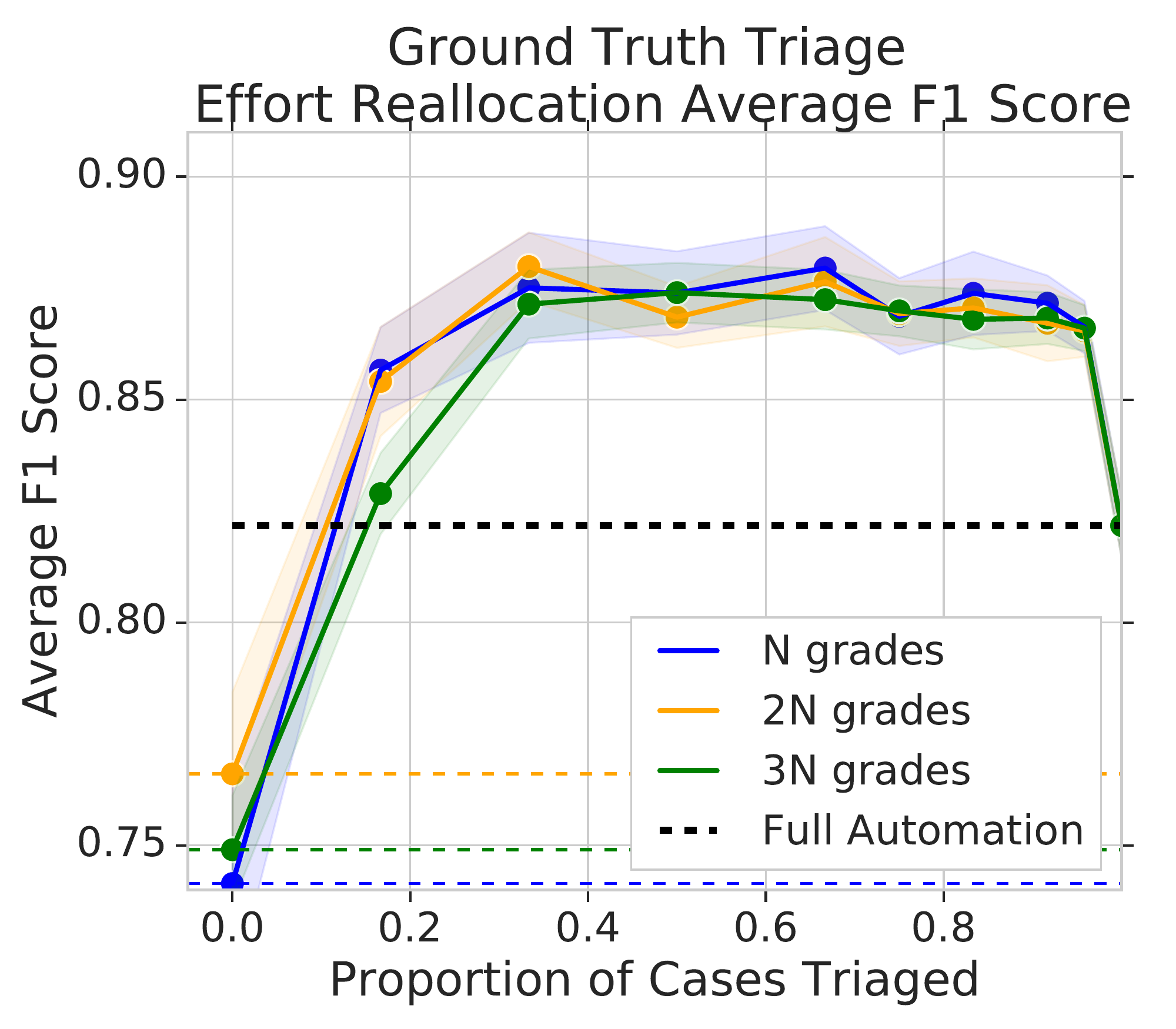} 
    \hspace*{3mm} \includegraphics[width=0.25\linewidth]{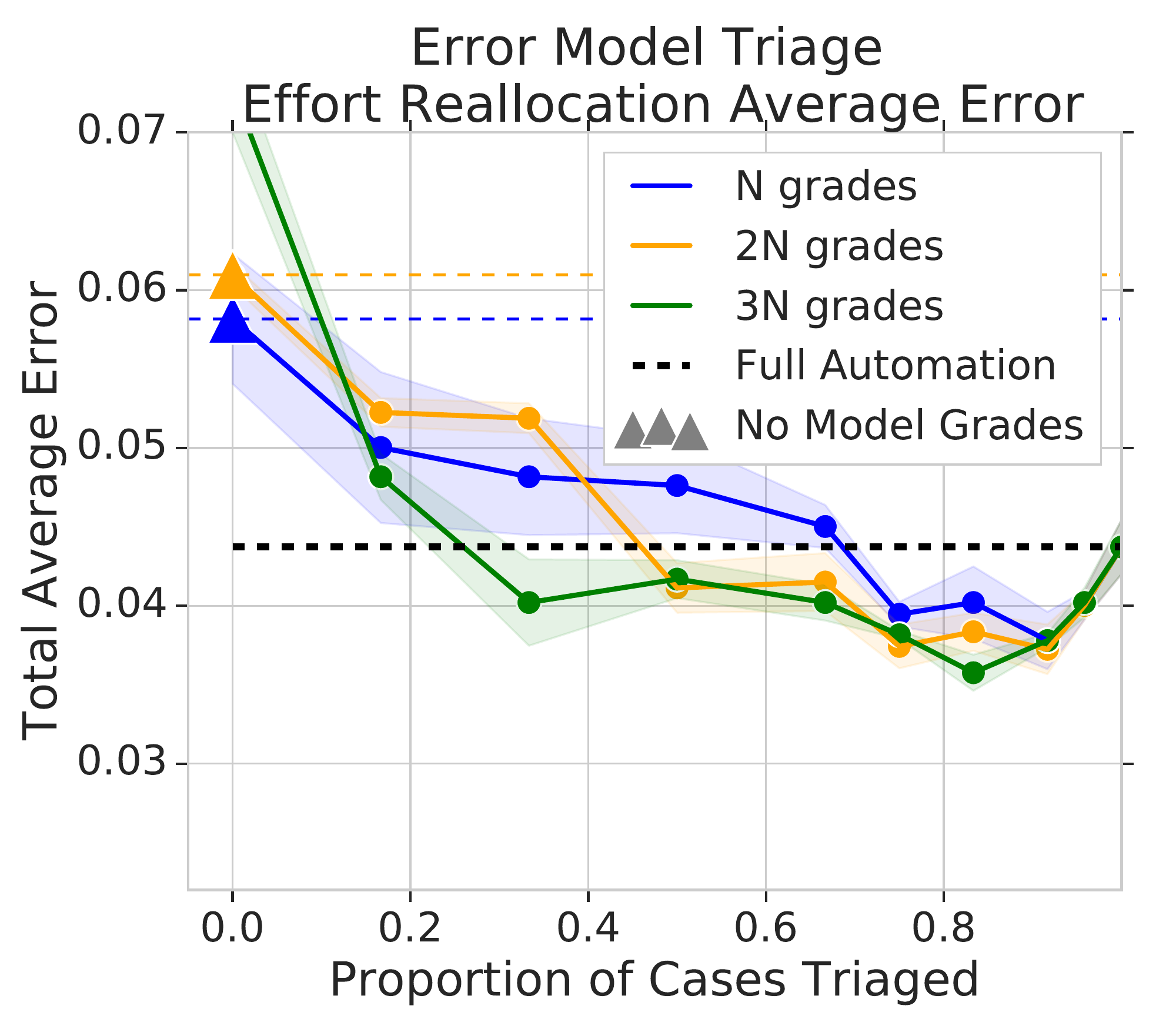} 
    &
    \hspace*{-6mm} \includegraphics[width=0.25\linewidth]{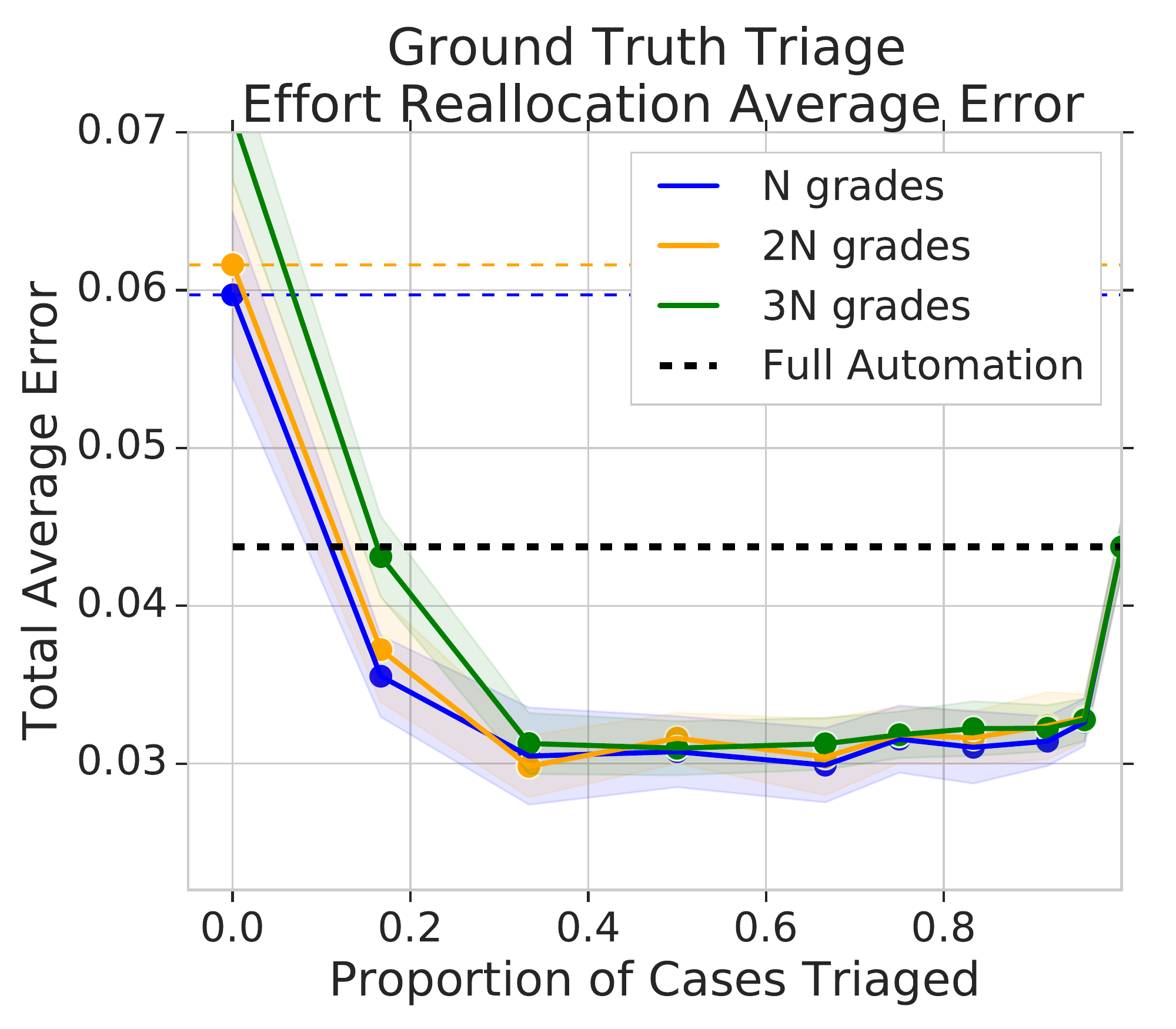} \\
    \hspace*{-10mm} \includegraphics[width=0.25\linewidth]{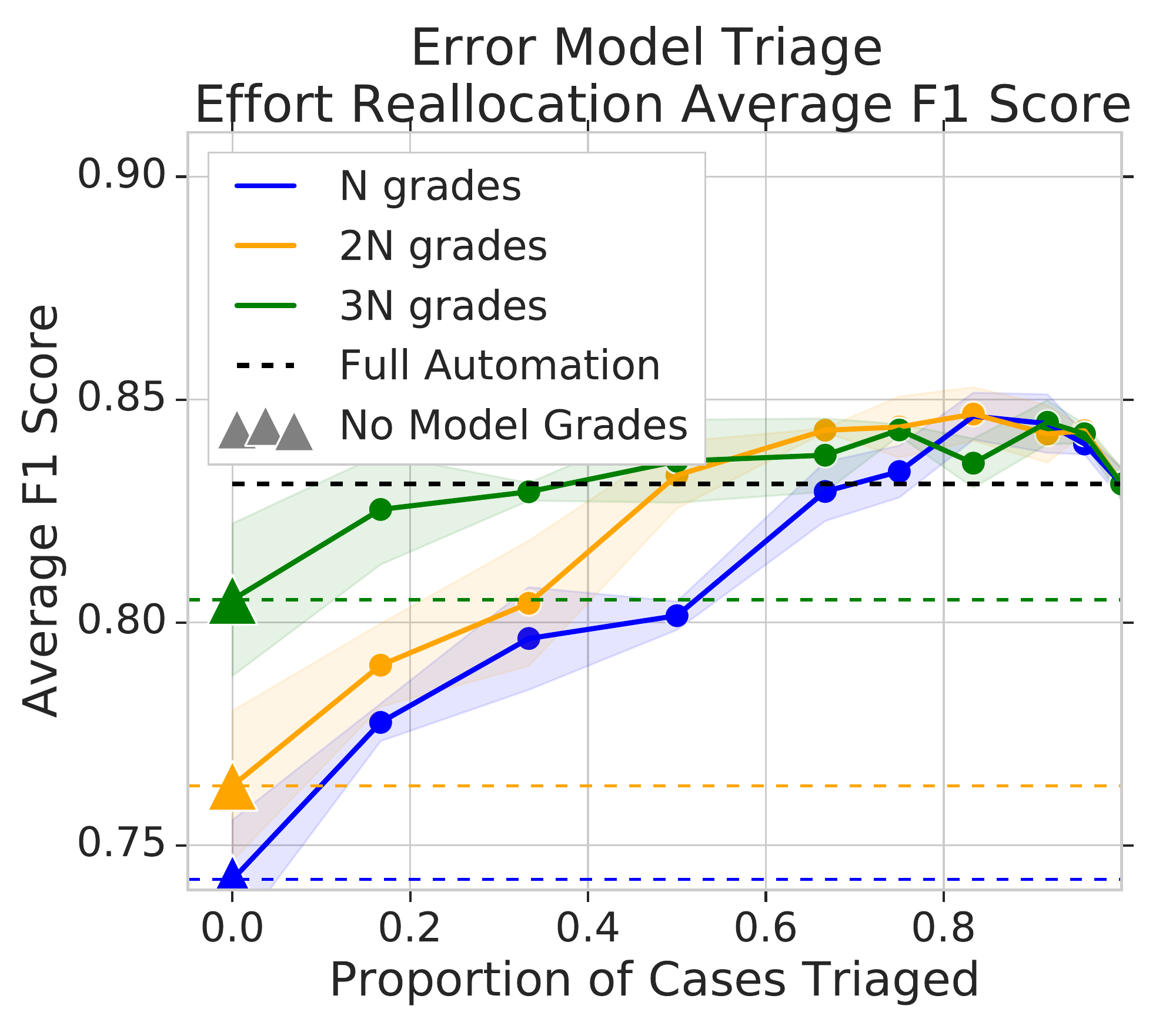} 
    &
    \hspace*{-6mm} \includegraphics[width=0.25\linewidth]{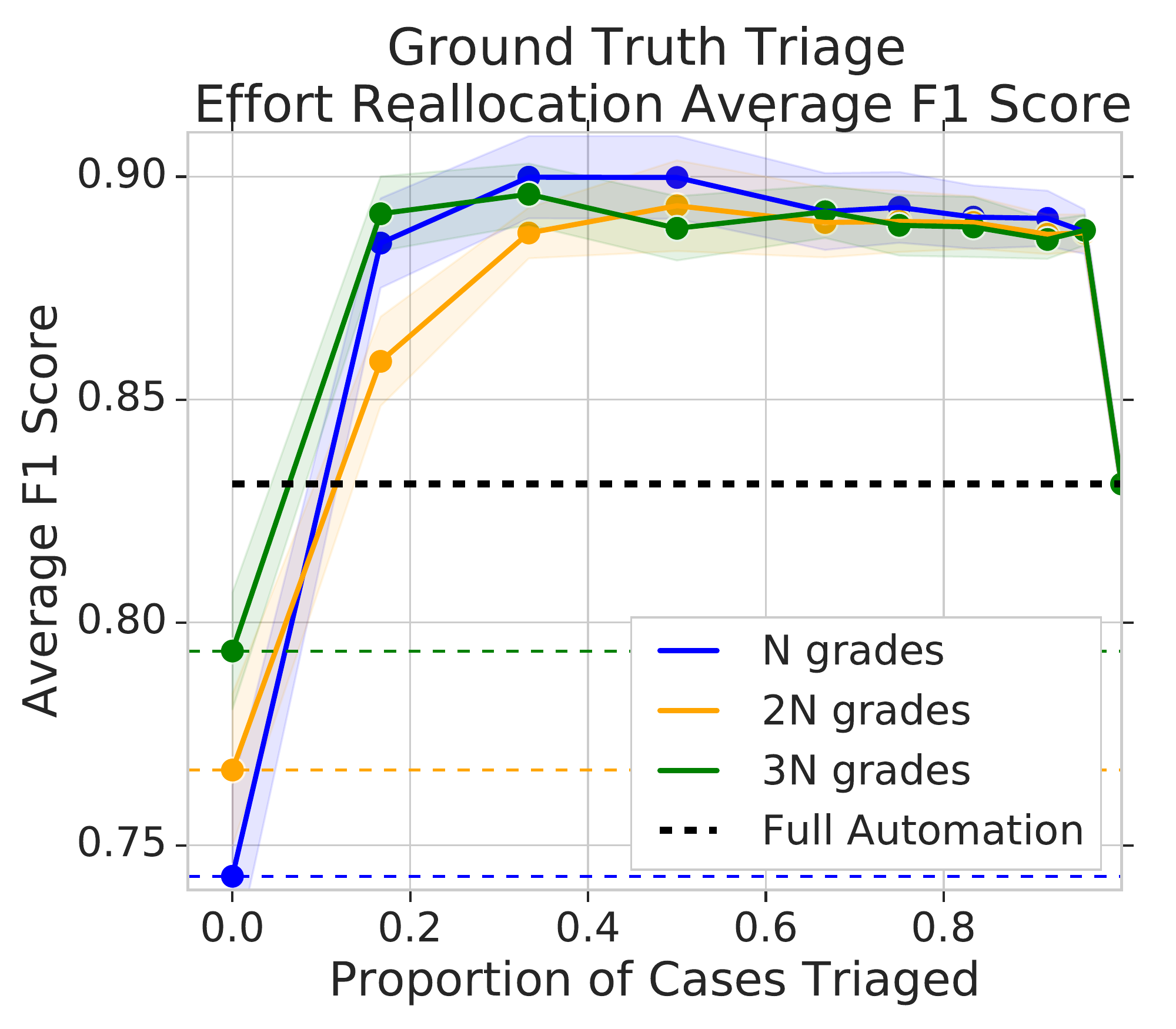} 
    \hspace*{3mm} \includegraphics[width=0.25\linewidth]{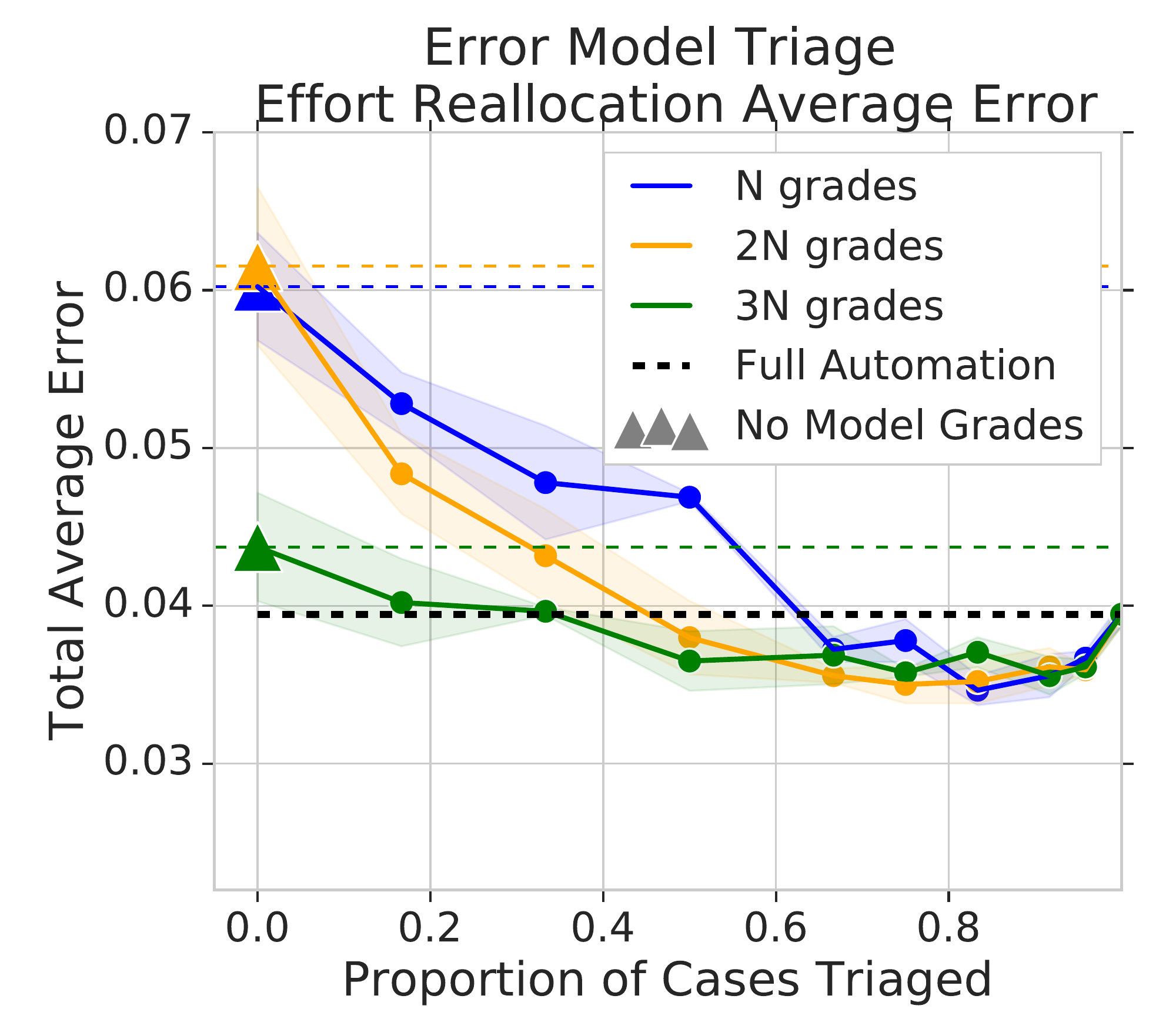} 
    &
    \hspace*{-6mm} \includegraphics[width=0.25\linewidth]{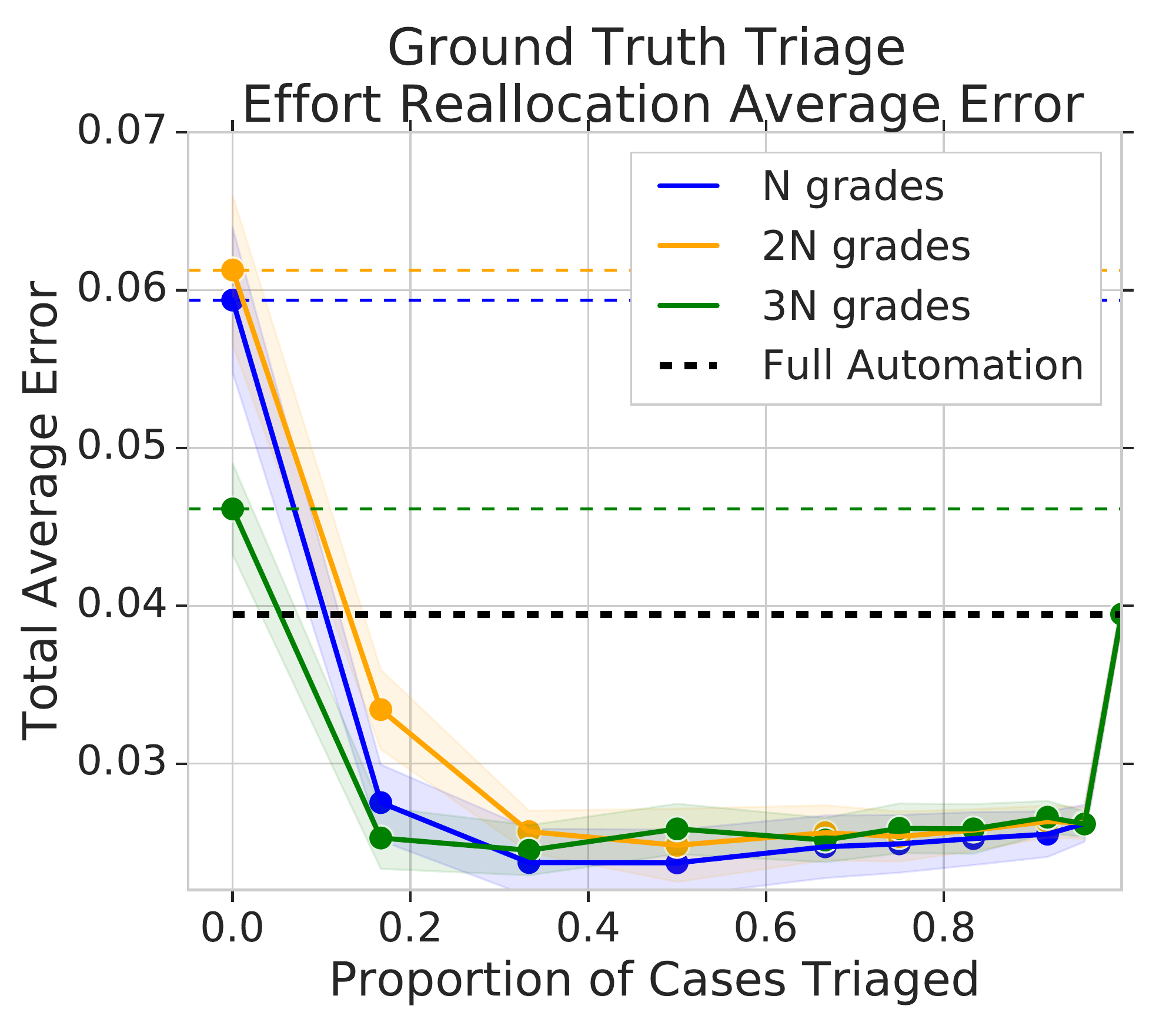}
    \end{tabular}
 \caption{\small \textbf{Triage and human expert effort reallocation for different thresholds: $0.3$ top row, $0.4$ bottom row.} We plot the same results as in Figure \ref{fig-triage-effort-reallocation} in the main text, but for different thresholds for aggregation (Section \ref{sec-aggregation}). In the main text, the threshold for referable after aggregating (averaging) the grades was $0.5$ (a majority vote), and here we show the results if the threshold used was $0.3$ (top row) or $0.4$ (bottom row). We see that the same qualitative conclusions hold as in Figure \ref{fig-triage-effort-reallocation}. }
 \label{fig-newthresh-effort-realloc}
\end{figure}

As described in Section \ref{sec-aggregation}, at evaluation, for an instance with multiple grades we aggregate all the scores by taking the mean and thresholding. In the main text, we pick this threshold to be $0.5$, corresponding to the majority vote of all the grades. In Figure \ref{fig-newthresh-effort-realloc}, we show the results corresponding to Figure \ref{fig-triage-effort-reallocation} in the main text, but for when we take the thresholds to be $0.3$ (top row) and $0.4$ (bottom row). We see that the qualitative conclusions remain the same -- combining human and algorithmic effort beats the full allocation and equal coverage protocols for both triage by the error prediction models and triage by the ground truth. We also see the same significant gap between ground truth and triage by error predictions. 

\begin{figure}
  \centering
  \begin{tabular}{cccc}
    \hspace*{-10mm} \includegraphics[width=0.25\linewidth]{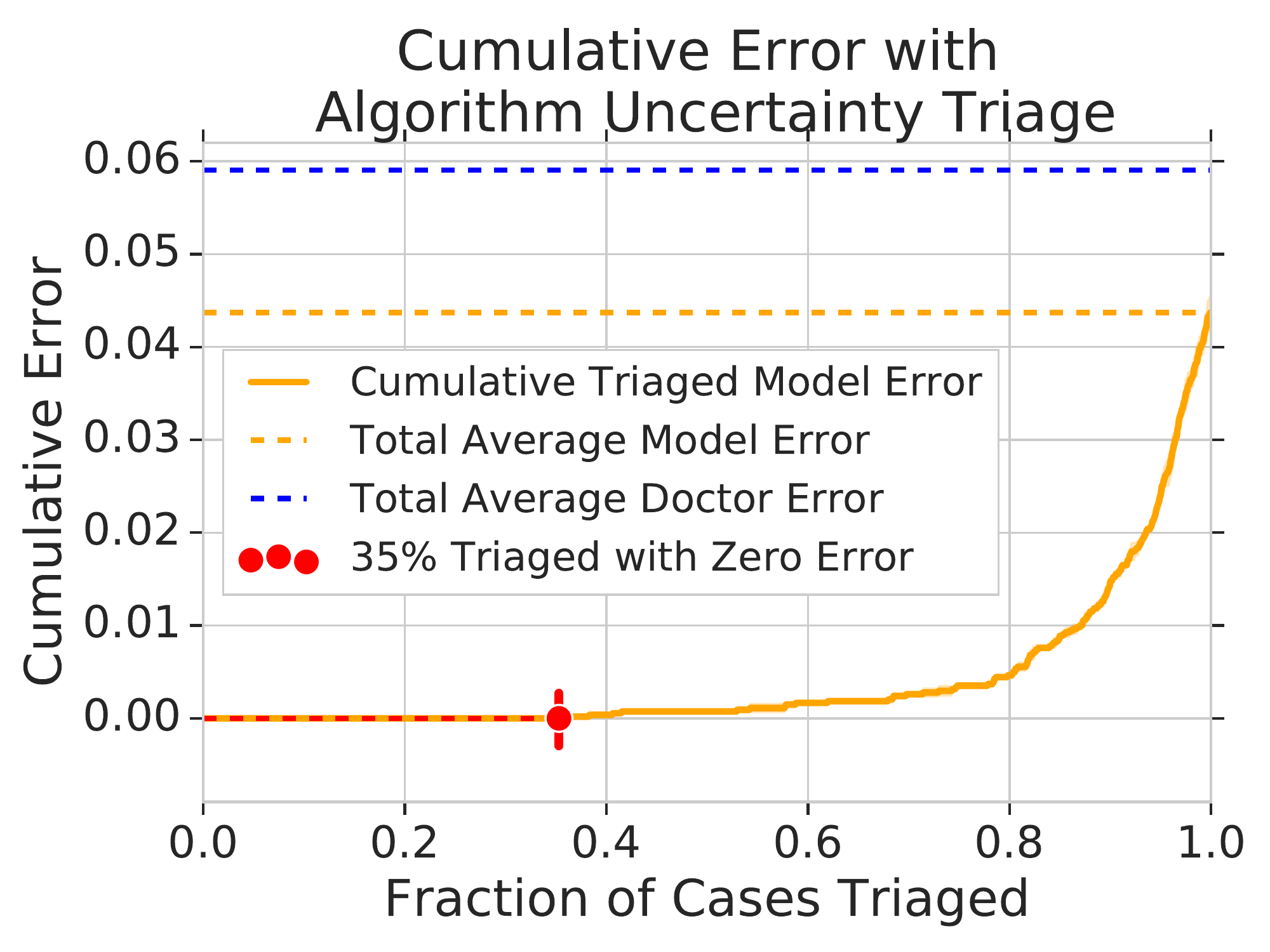}
  &
\hspace*{-5mm} \includegraphics[width=0.25\linewidth]{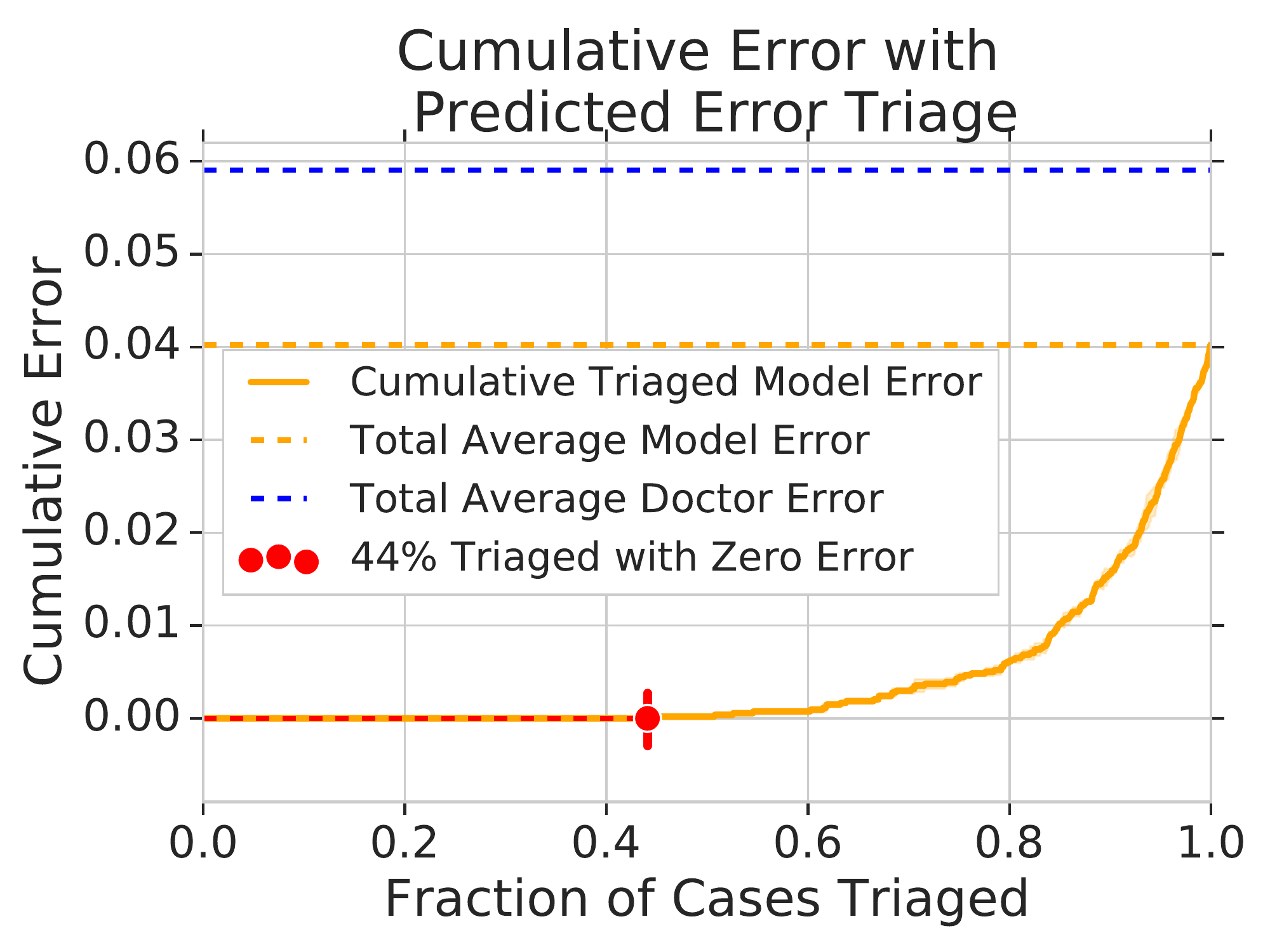} &
\hspace*{2mm} \includegraphics[width=0.25\linewidth]{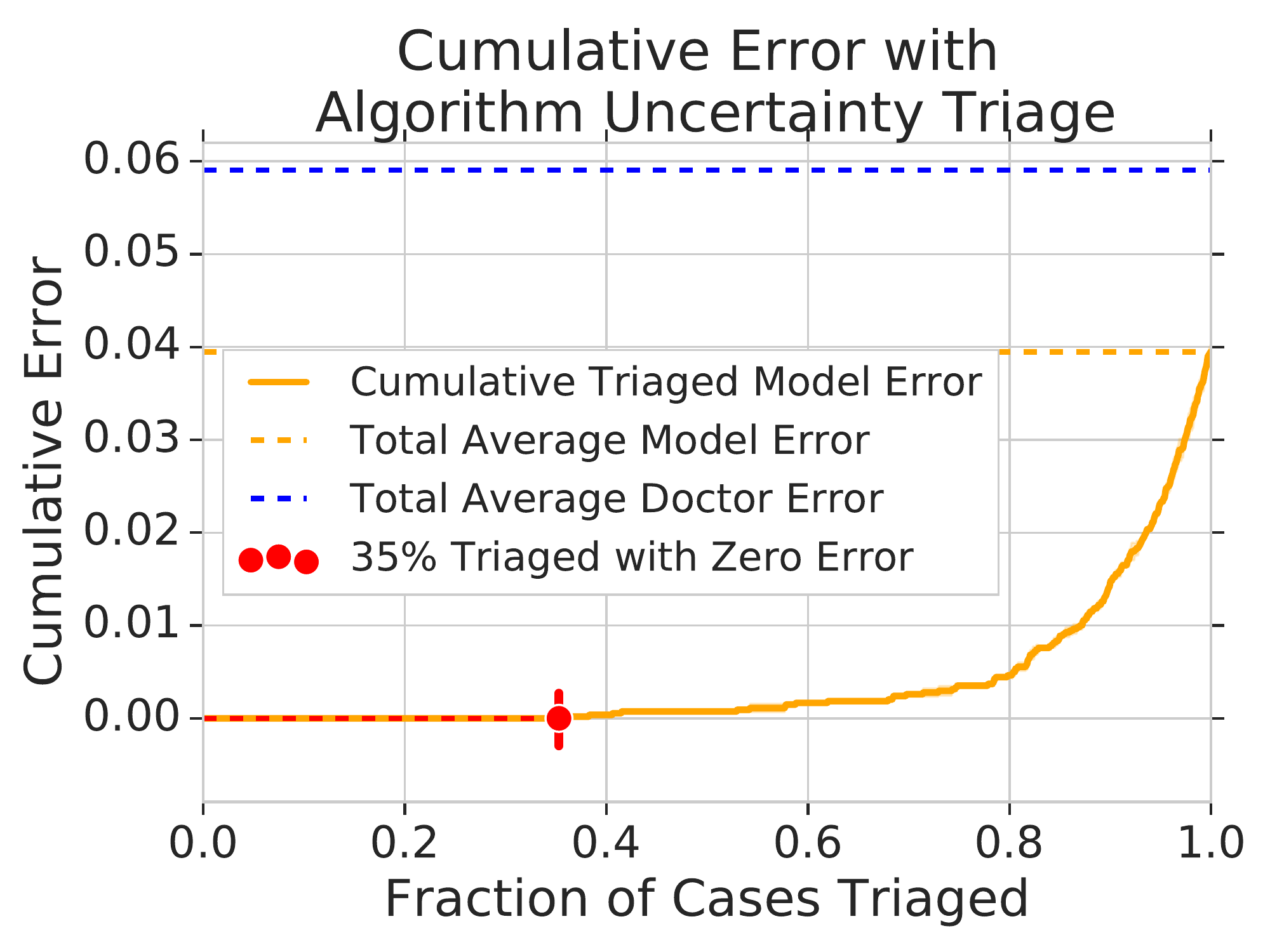}
  &
\hspace*{-5mm} \includegraphics[width=0.25\linewidth]{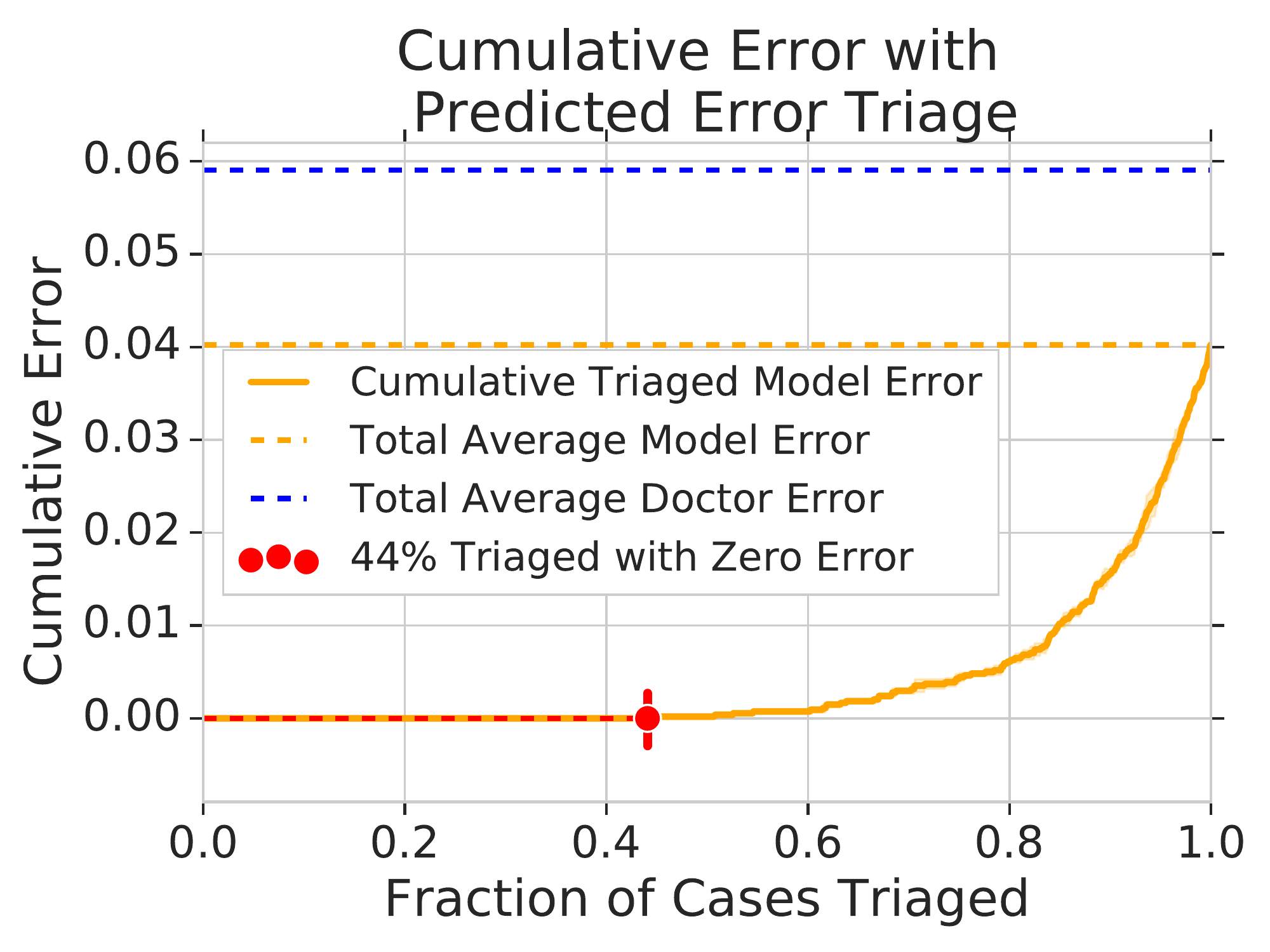}
 \end{tabular}
 \caption{\small \textbf{Triage for Zero Error subsets with thresholds $0.3$ (left) and $0.4$ (right) for aggregation.} The size of the zero error subsets remain the same, showing that the choice of threshold does not affect the identification of these subsets.}
 \label{fig-newthresh-zeroerr}
\end{figure}

Note that this choice of threshold affects the choice of $q_R$, which is chosen so that the number of cases marked as referable by the model matches the number of cases marked as referable by the aggregated and thresholded grade of the human doctors, and could potentially affect the results of Figure \ref{fig-effort-saving}. However, as shown in Figure \ref{fig-newthresh-zeroerr}, the choice of aggregation threshold does not affect the identification of zero error subsets.

\section{Triage and Human Effort Reallocation with Model Grades}
\label{app-sec-alg-grades}
\begin{figure}
  \centering
  \begin{tabular}{cccc}
    \hspace*{-10mm} \includegraphics[width=0.25\linewidth]{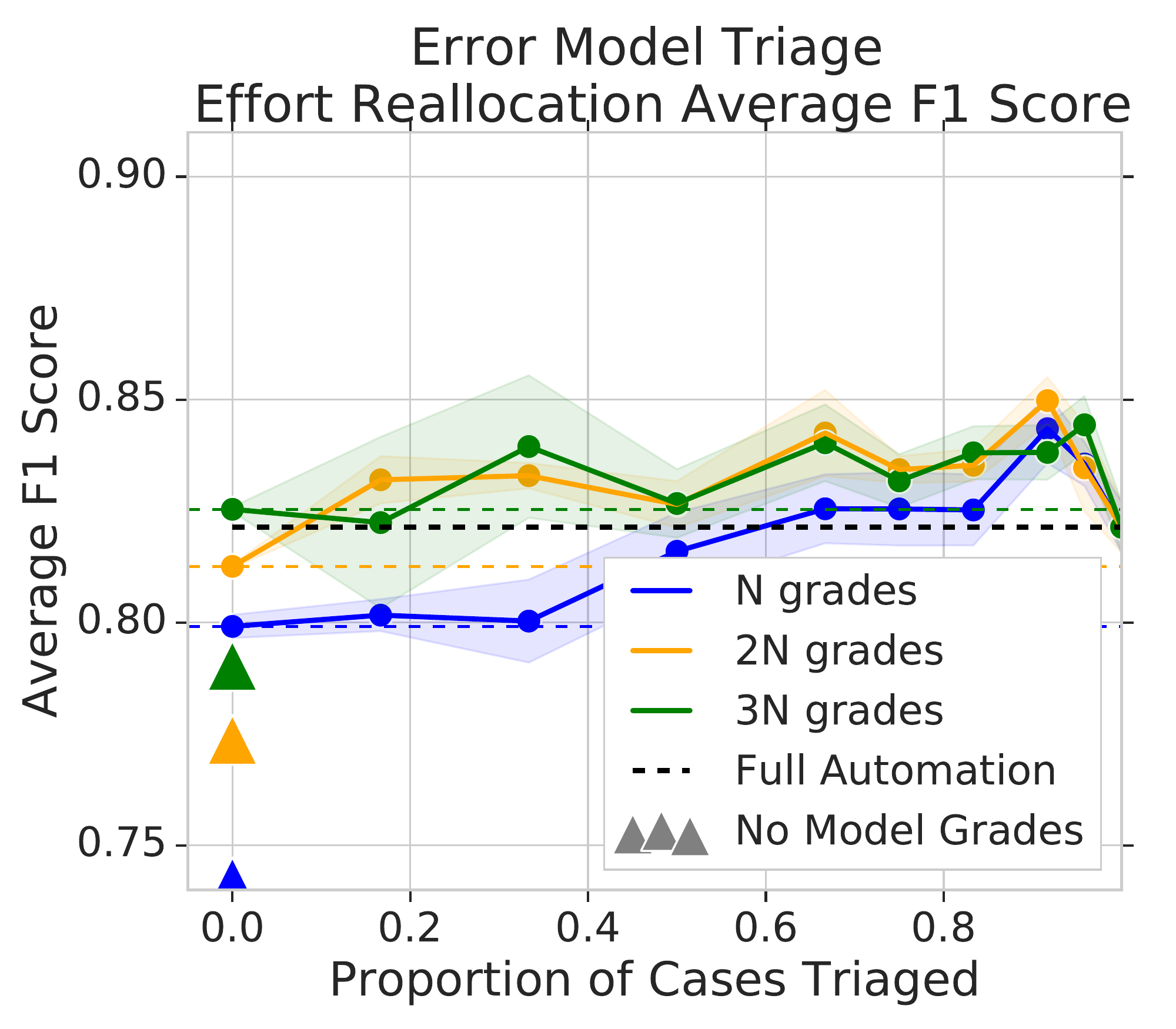} 
    &
    \hspace*{-6mm} \includegraphics[width=0.25\linewidth]{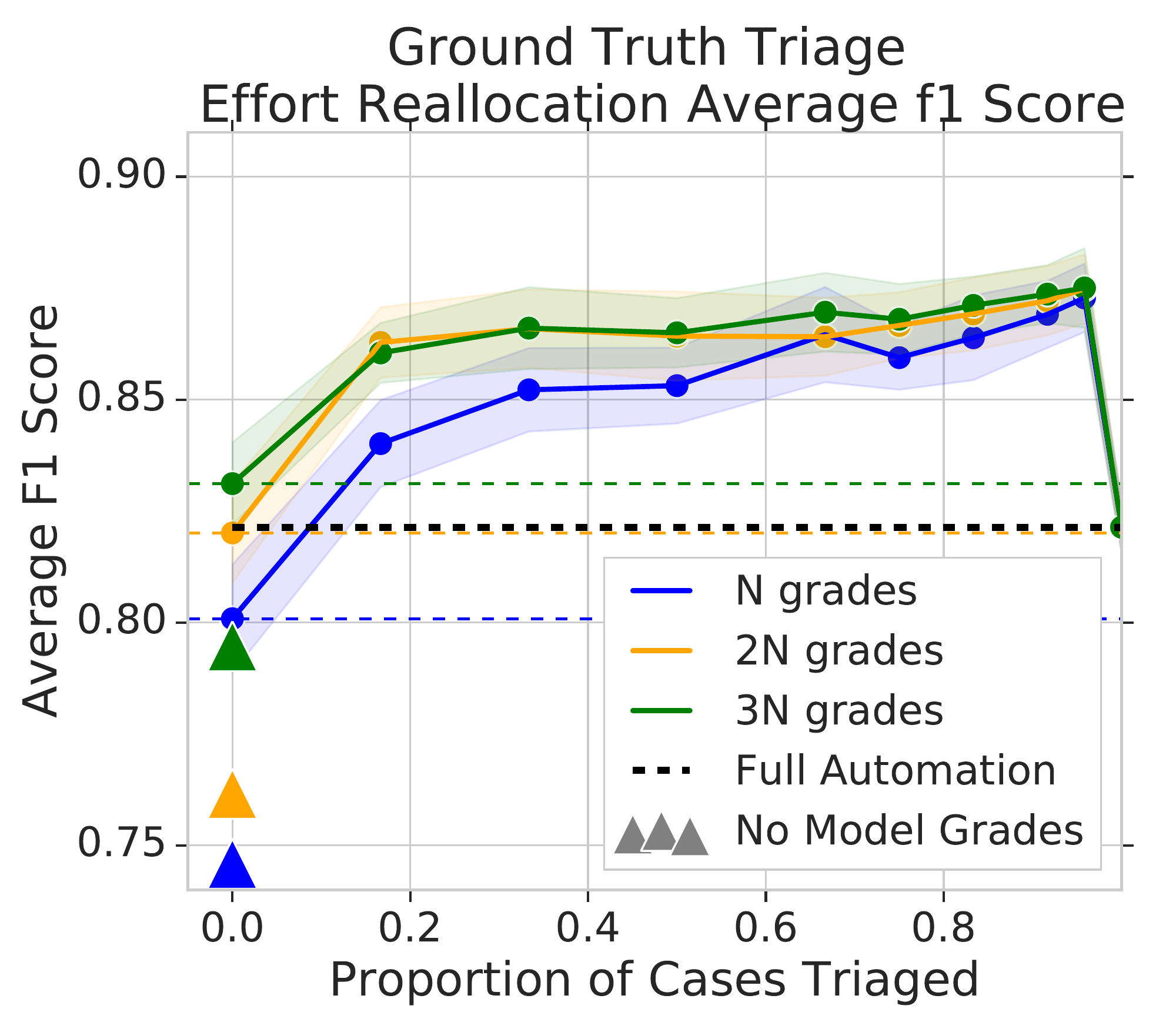} 
    \hspace*{3mm} \includegraphics[width=0.25\linewidth]{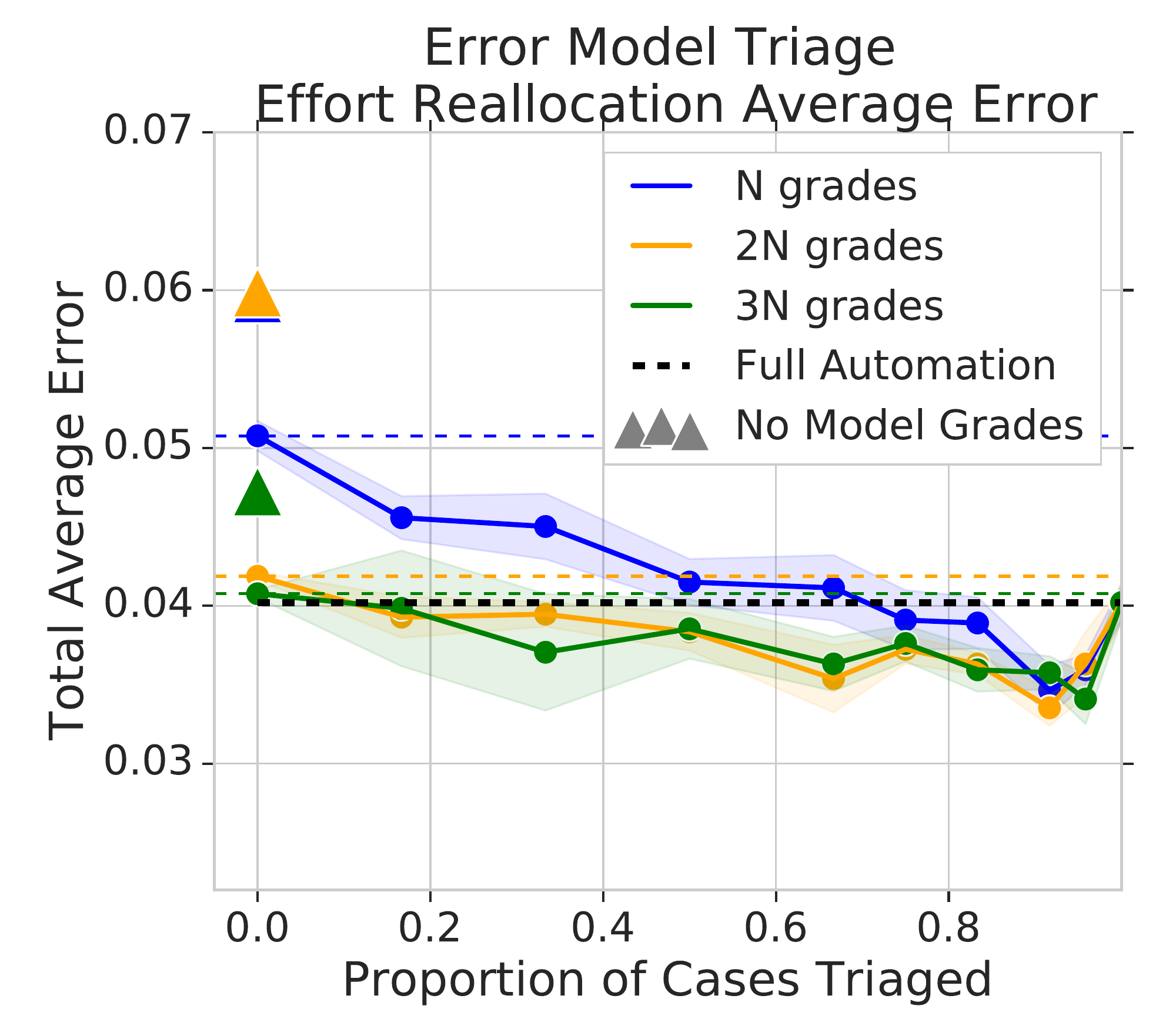} 
    &
    \hspace*{-6mm} \includegraphics[width=0.25\linewidth]{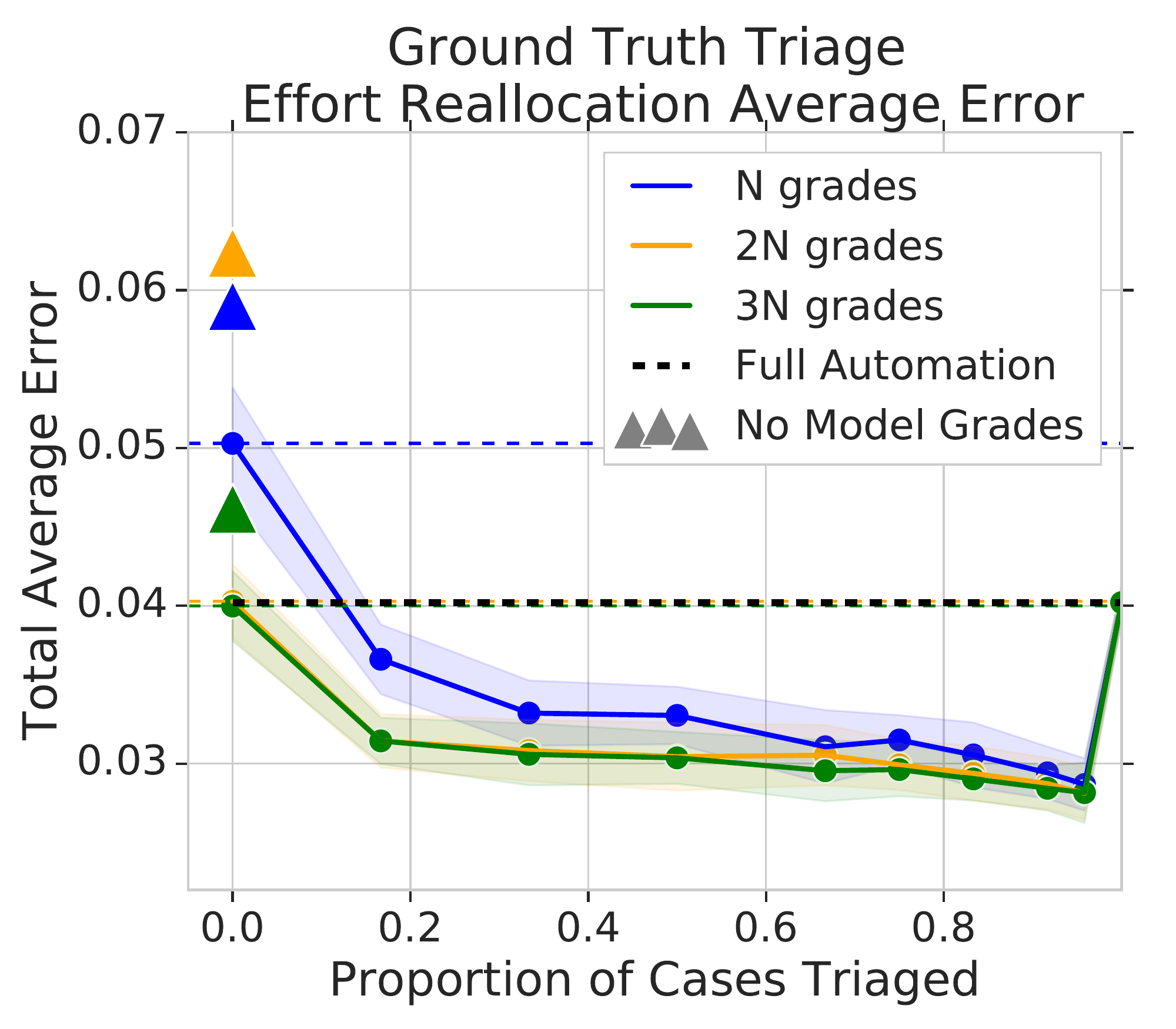}  
    \end{tabular}
 \caption{\small \textbf{Effort reallocation results with the algorithm's grade being available for all cases.} Here we assume that the algorithm's grade is available for all the patient cases. We see that the same qualitative conclusions hold -- a mix of automation and human effort outperforming the pure algorithm/human expert.}
 \label{fig-app-model-grades}
\end{figure}

The triage process for effort reallocation -- Figures \ref{fig-triage-effort-reallocation}, \ref{fig-triage-effort-reallocation-eyepacs} -- assumes that the algorithm decision is not available for the $(1 - \alpha)N$ cases that are not automated. This may be the situation if computing an algorithm decision is expensive (less likely) or (more likely) the algorithm decision is purposefully not shown in cases where it is unsure, so as not to bias the human doctors. However, another equally likely scenario is that the algorithm decision is also available `for free' for the $(1 - \alpha)N$ cases that are not fully automated. In Figure \ref{fig-app-model-grades}, we show the effort reallocation results from triaging if the model grades were available for all the cases (compare to Figure \ref{fig-triage-effort-reallocation} in the main text). We observe that all of the main conclusions -- the optimal performance is through a combination of automation and human effort, which beats both full automation and the different equal coverage baselines. 

\section{Results on Additional Holdout Dataset}
\label{app-sec-other-dataset}
The results in the main text are on the adjudicated evaluation dataset, which, aside from multiple independent grades by individual doctors, also have a consensus score, the adjudicated grade, which is used as a proxy for ground truth. To further validate our result, we use an additional holdout set which doesn't have an adjudicated grade, but does have many individual doctor grades. For each instance $i$, we use half of its grades to compute a proxy ground truth grade, by aggregating and then thresholding the doctor grades. The other half of the grades are used in effort reallocation and evaluating the equal coverage baseline. The individual doctor grades in this dataset are slightly noiser (higher disagreement rates) than in the adjudicated evaluation dataset. Nevertheless this additional evaluation also supports all of the main findings.

\begin{figure}
  \centering
      \begin{tabular}{cc}
    \hspace*{-10mm} \includegraphics[width=0.4\linewidth]{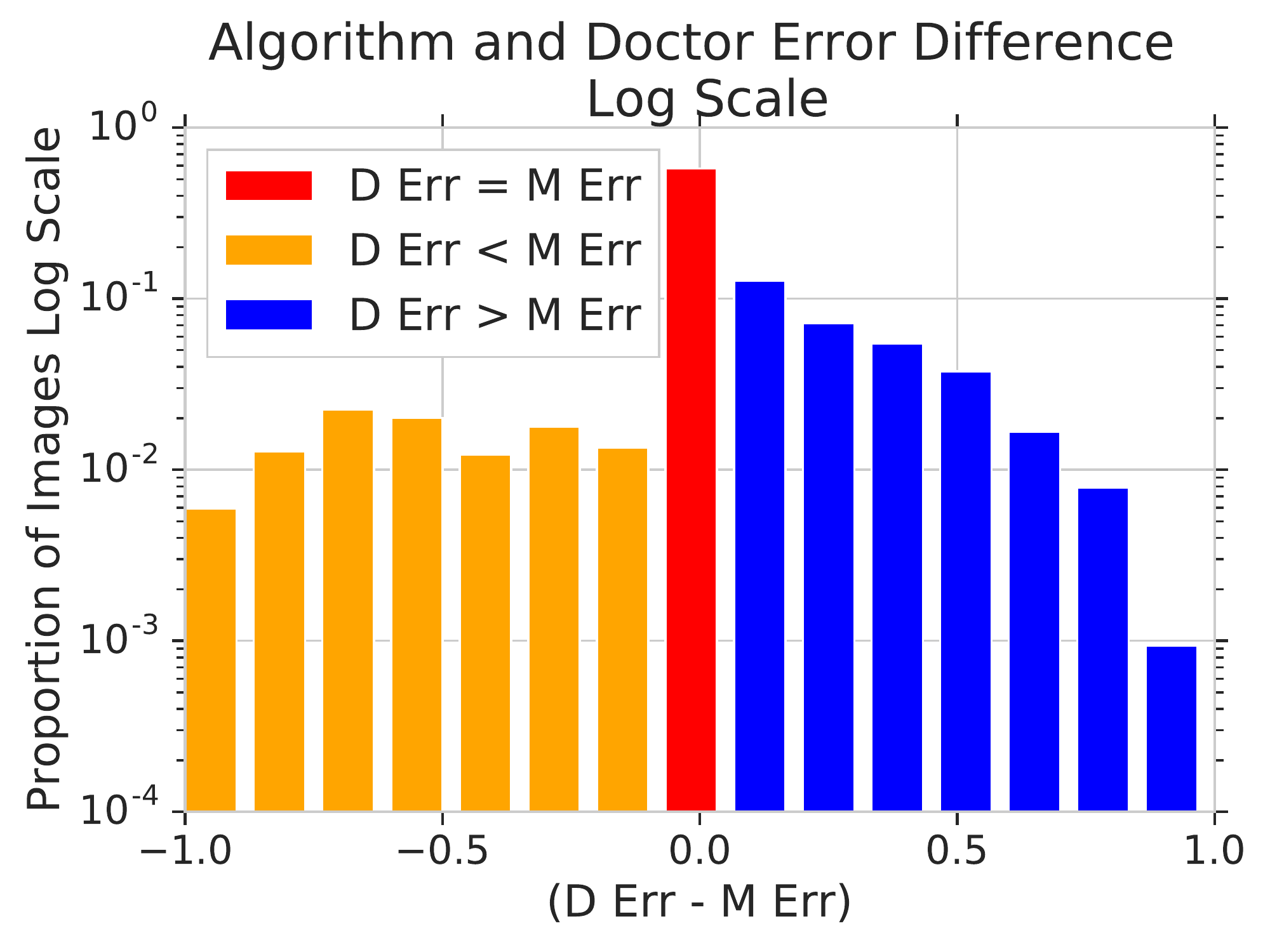} 
    &
    \hspace*{-2mm} \includegraphics[width=0.4\linewidth]{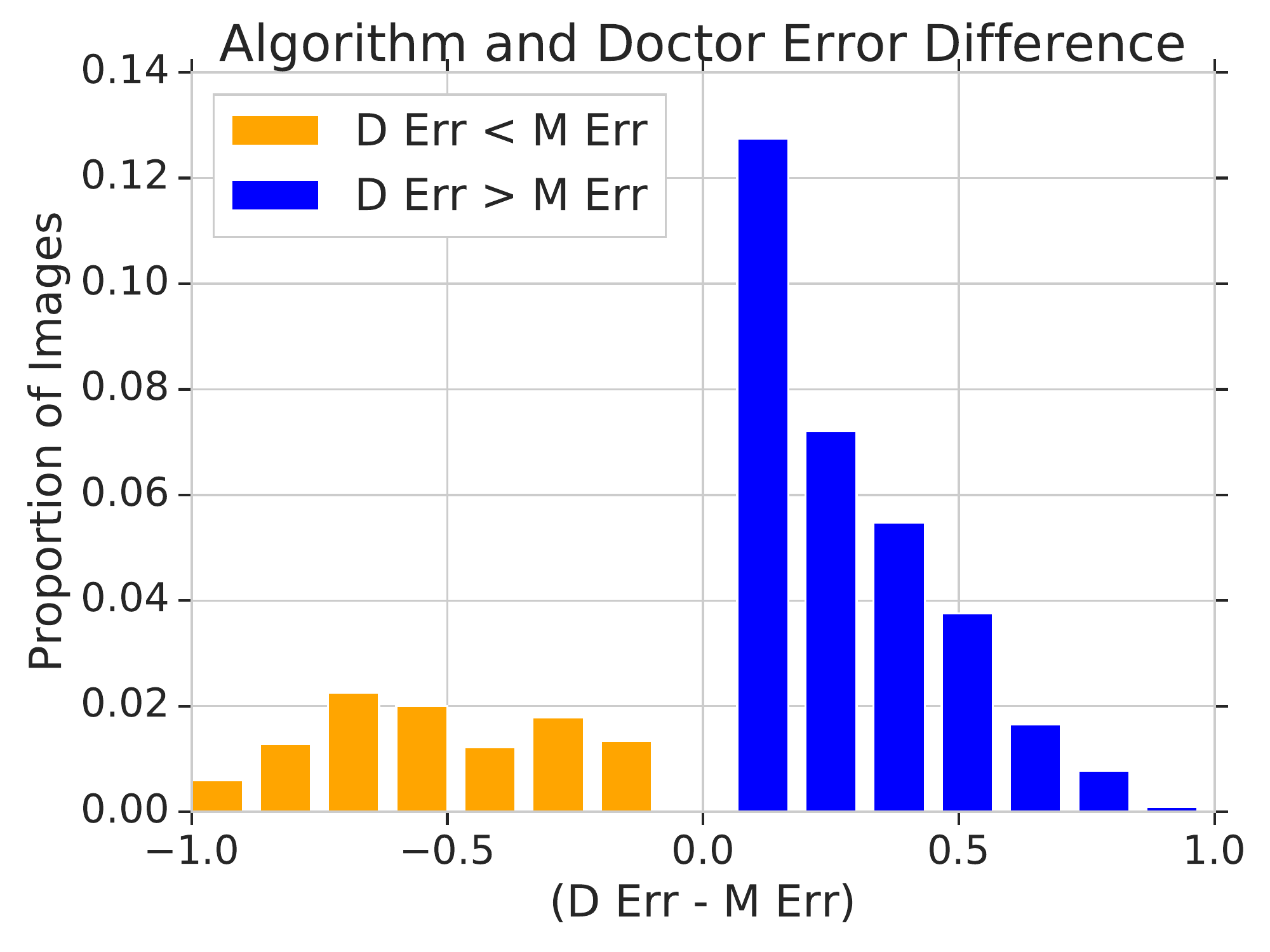} \\
    \end{tabular}
 \caption{\small \textbf{Histogram plot of $\Prb{H_i} - \Prb{M_i}$ for instances $i$ on the additional holdout evaluation dataset.} Compare to Figure \ref{fig-scatterplot-adj} in the main text. We see a diversity of values across different instances.}
 \label{fig-scatterplot-eyepacs}
\end{figure}

The results on this dataset are qualitatively identical to those with the adjudicated data. Again, we see that there is a diverse spread of $\Prb{H_i} - \Prb{M_i}$ across instances, with around $10\%$ of the instances having the human experts perform better (Figure \ref{fig-scatterplot-eyepacs}).

\begin{figure}
\vspace*{-10mm}
  \centering
  \begin{tabular}{cc}
    \hspace*{-5mm} \includegraphics[width=0.5\linewidth]{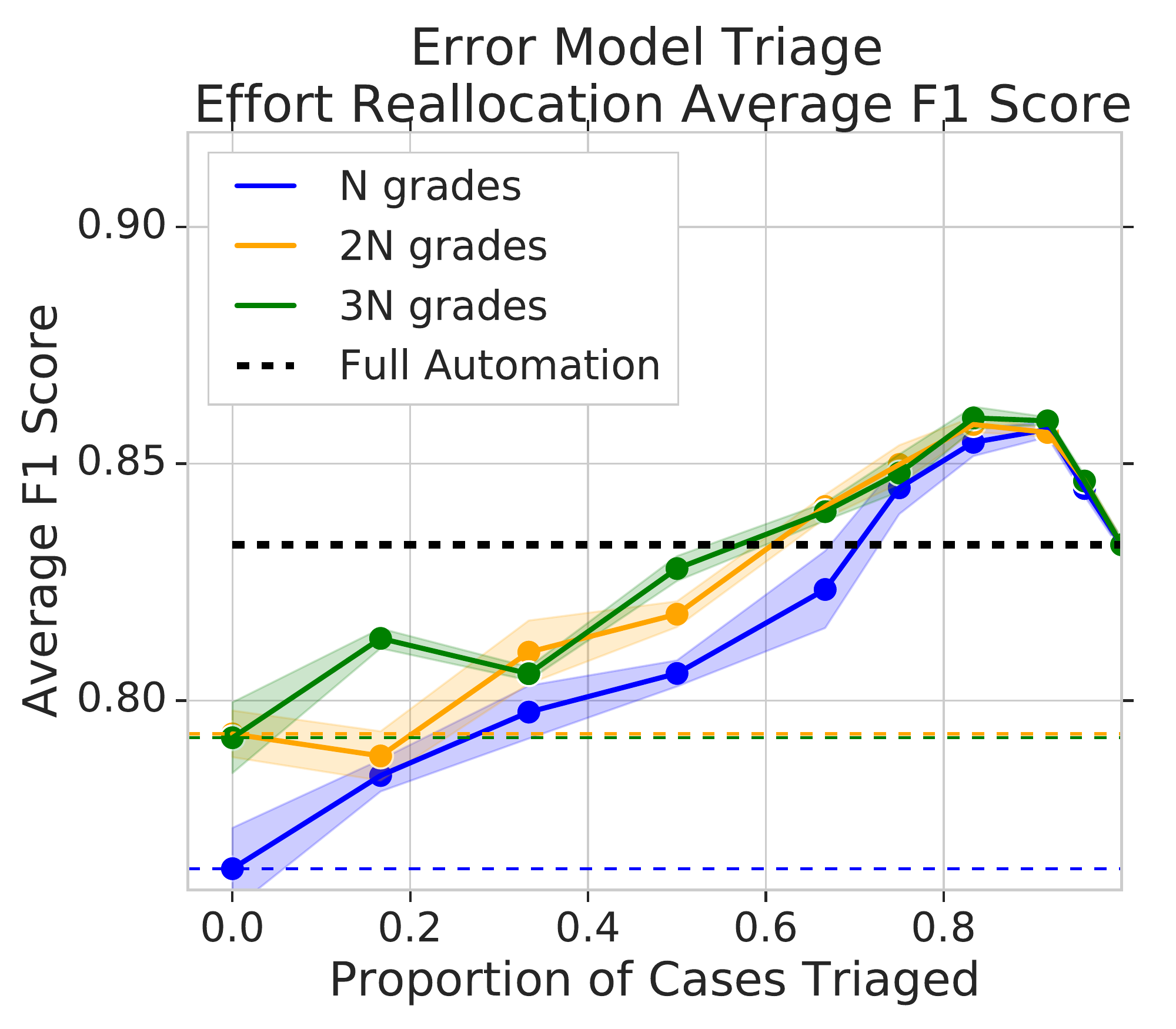} 
    &
    \hspace*{-3mm} \includegraphics[width=0.5\linewidth]{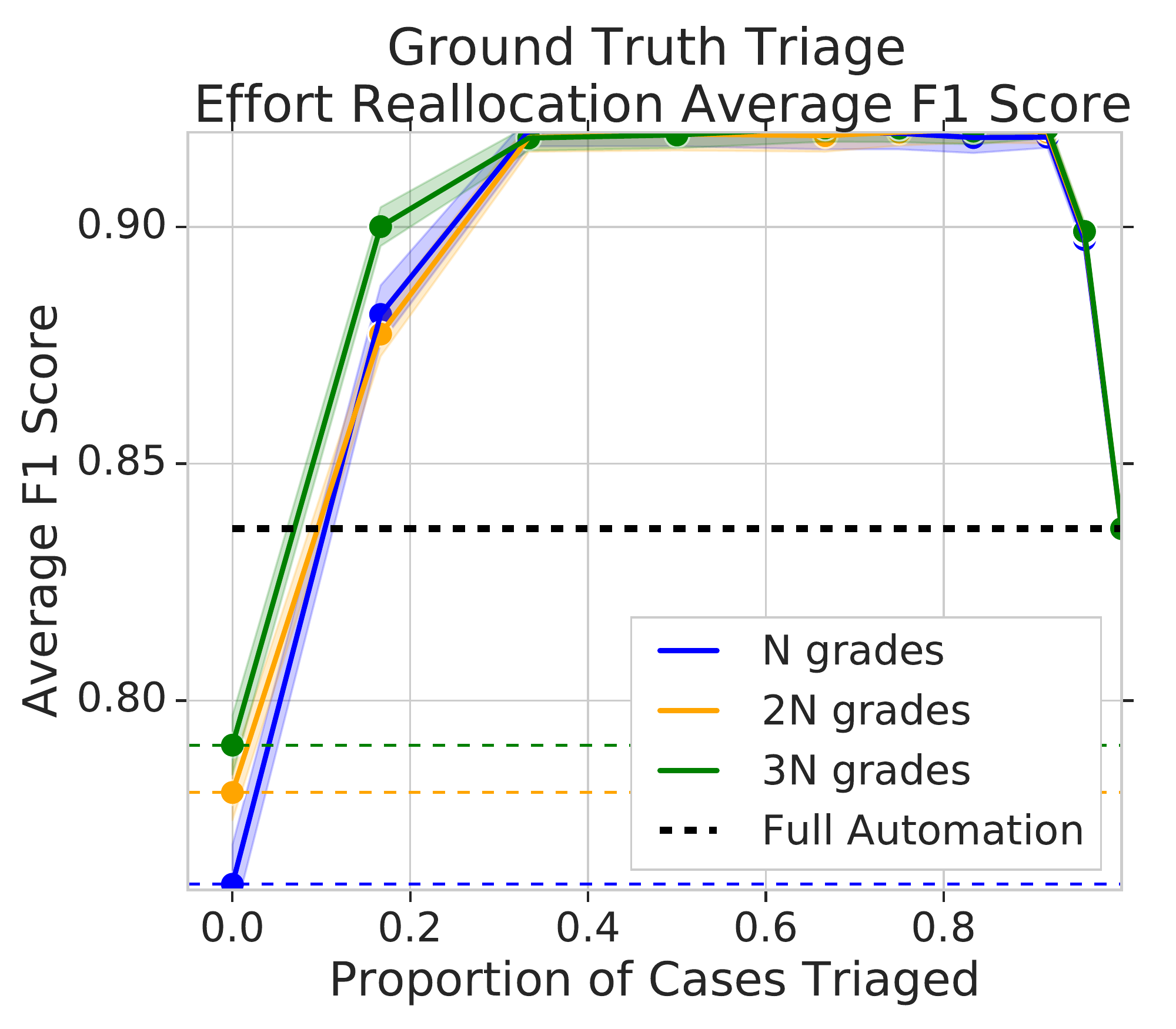} \\
    \hspace*{-5mm} \includegraphics[width=0.5\linewidth]{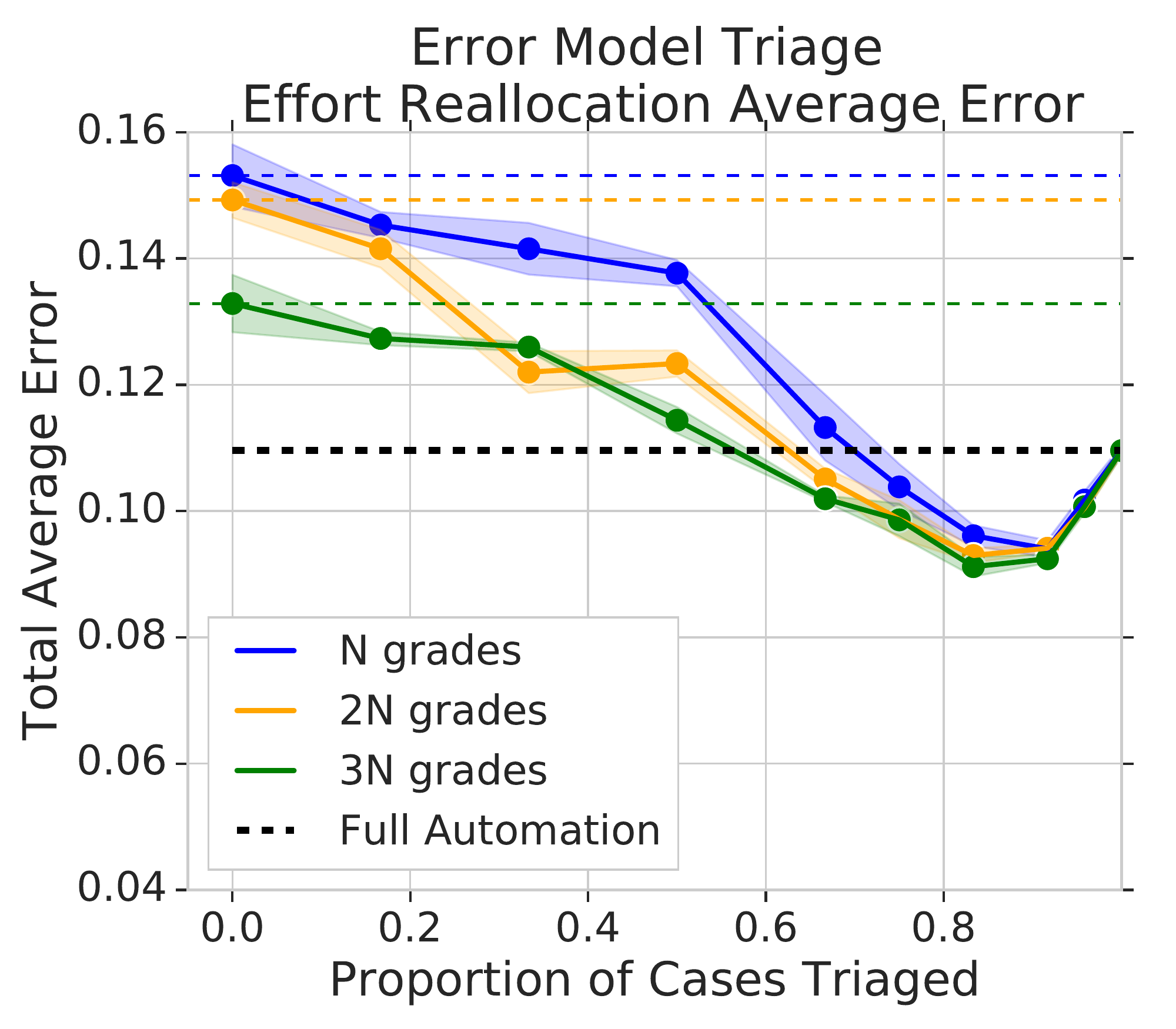} 
    &
    \hspace*{-3mm} \includegraphics[width=0.5\linewidth]{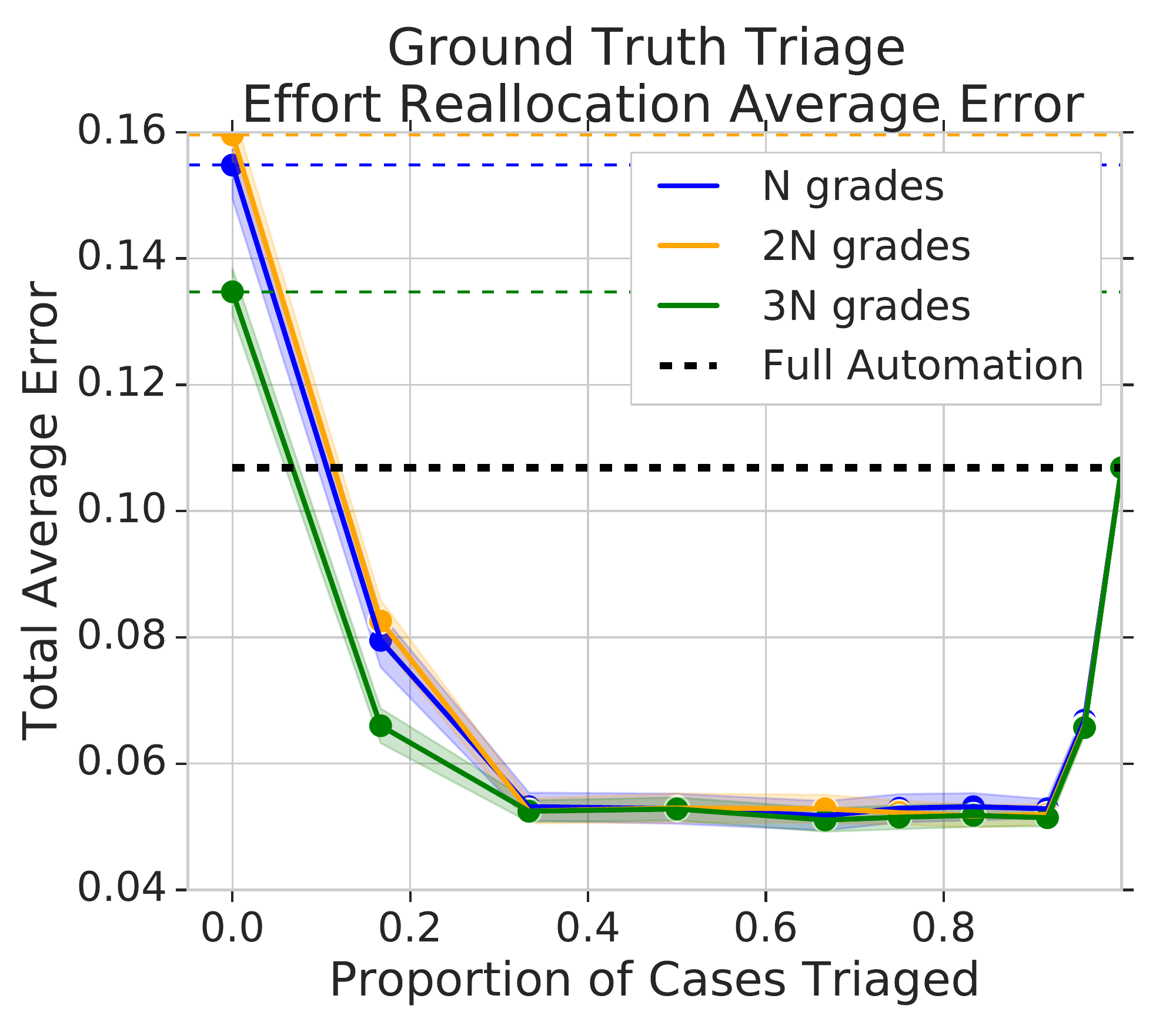}
    \end{tabular}
 \caption{\small \textbf{Triaging to combine human effort and algorithm decisions outperforms full automation and the equal coverage protocols, compare to Figure \ref{fig-triage-effort-reallocation} in the main text.} We perform the same experiments as in Figure \ref{fig-triage-effort-reallocation} in the main text, using the aggregated doctor grades as the ground truth instead of the adjudicated grade.}
 \label{fig-triage-effort-reallocation-eyepacs}
\end{figure}

This diversity continues to be predictable, and we observe that triaging (by the error prediction models and by the ground truth) to combine human expert effort and and the algorithm's decisions, Figure \ref{fig-triage-effort-reallocation-eyepacs}, also demonstrates that this combination works better than both full automation and equal coverage -- the same conclusions seen in Figure \ref{fig-triage-effort-reallocation} in the main text. Like the main text, we see a gap between triaging by the error prediction models and the ground truth score. 

Finally, we also test to see if triaging can help find sets of zero error, like in Section \ref{sec-automation} and Figure \ref{fig-effort-saving}. We find that this is indeed the case, though the fractions are slightly smaller with this holdout dataset, likely because the labels are noisier than on the adjudicated evaluation dataset.

\begin{figure}
  \centering
  \begin{tabular}{cc}
    \hspace*{-10mm} \includegraphics[width=0.5\linewidth]{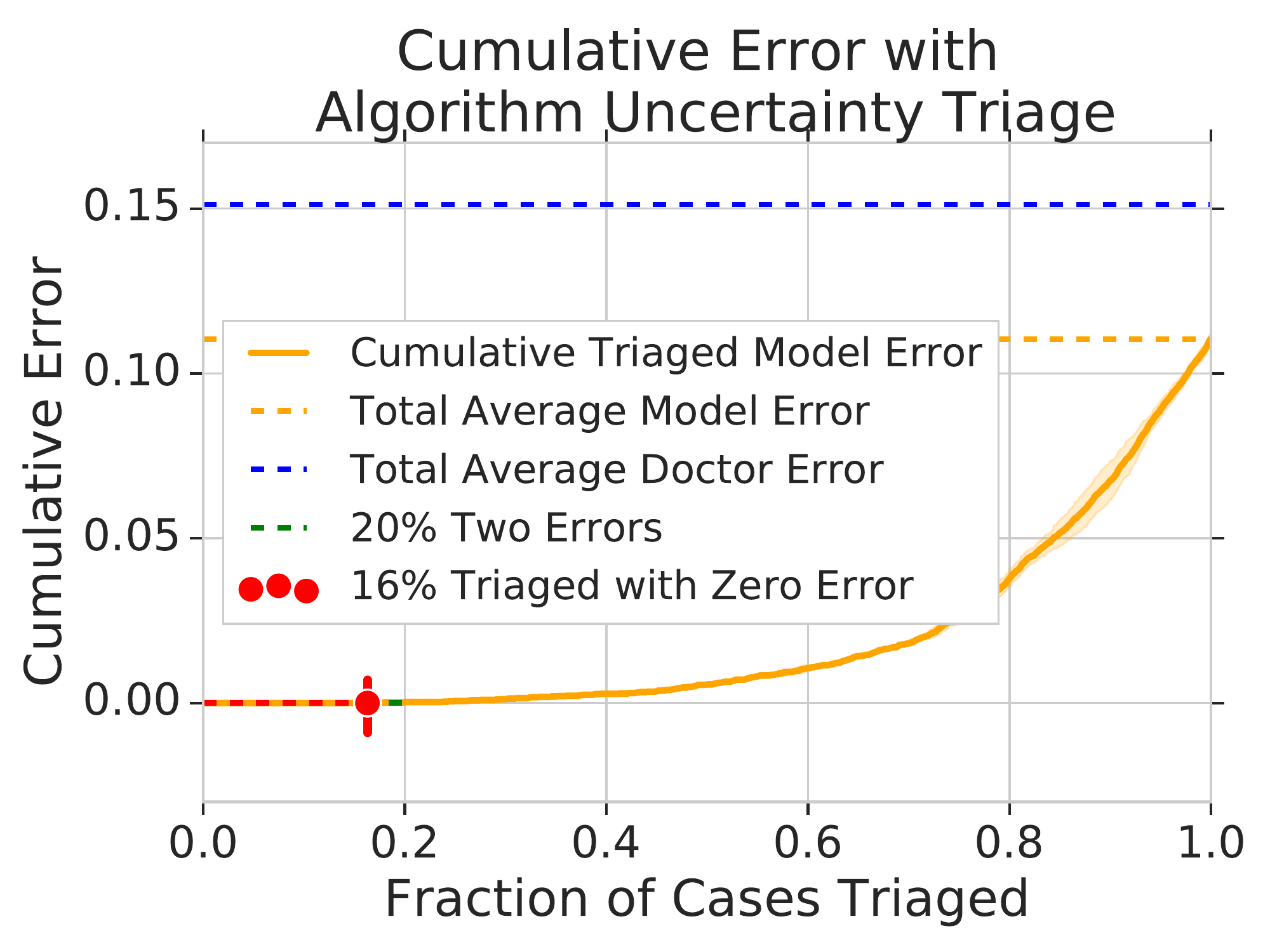}
  &
\includegraphics[width=0.5\linewidth]{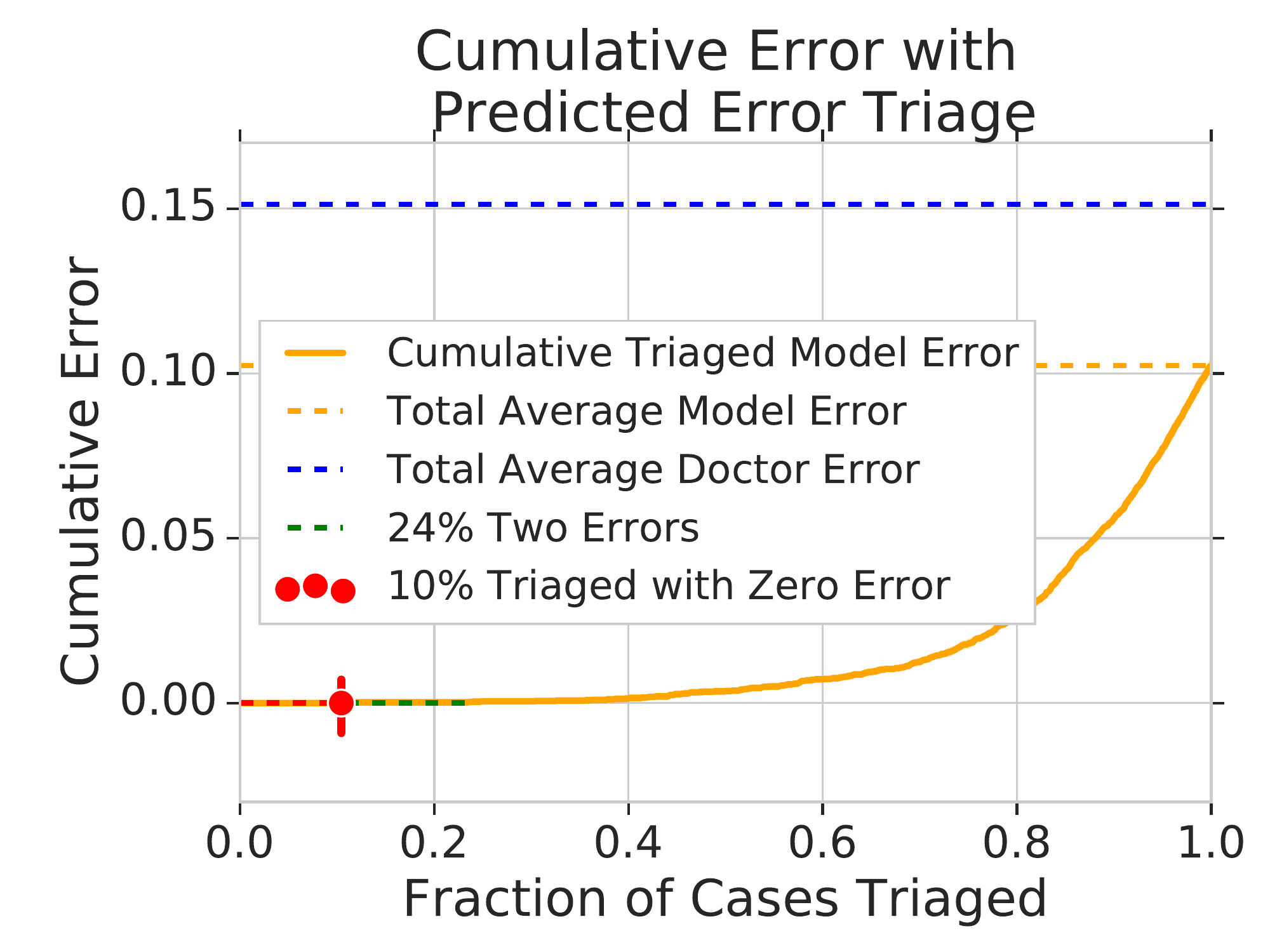} 
 \end{tabular}
 \caption{\small \textbf{Proportion of data with zero errors when triaging.} Compare to Figure \ref{fig-effort-saving} in the main text. The proportion of the dataset found with zero errors is slightly lower, likely because the labels are noisier, and total error is much higher ($10\%$ compared to $4\%$ in the main text.) Unlike the main text, we see that triaging by algorithmic uncertainty, left pane, seems to perform better than triaging with a separate model. Upon closer inspection, we find that this is because one repetition of the separate error model makes two errors earlier on, and accounting for this (green dotted lines) shows that the separate error model performs comparably/slightly better.}
 \label{fig-effort-saving-eyepacs}
\end{figure}

We also see that the fraction of zero error examples triaged is slightly lower with the separate error prediction model (Figure \ref{fig-effort-saving-eyepacs} right) than triaging by model uncertainty (Figure \ref{fig-effort-saving-eyepacs} left). The reason for this becomes apparent after further inspection: the results of Figure \ref{fig-effort-saving-eyepacs} are averaged over three independent repetitions of training a main diagnostic model, and a corresponding separate error model. We find that one of the three repetitions of the separate error model makes two errors -- at $10\%$ of the way through the data, it triages two examples that are errors.This causes the percentage with zero error to drop from $24\%$ to $10\%$. If we account for these two errors, we see that in fact triaging by the error prediction model is doing comparably to triaging by algorithm uncertainty, where allowing two errors gets to $20\%$ of the data.

\label{sec-automation}

\end{document}